\documentclass[11pt]{aicosmos}

\usepackage{csquotes}
\usepackage{longtable}
\usepackage[bottom]{footmisc}

\usepackage[utf8]{inputenc} 
\usepackage[T1]{fontenc}    
\usepackage{hyperref}       
\usepackage{url}            
\usepackage{booktabs}       
\usepackage{amsfonts}       
\usepackage{nicefrac}       
\usepackage{microtype}      
\usepackage{xcolor}         

\usepackage{amsmath}
\usepackage{amssymb}
\usepackage{mathtools}
\usepackage{amsthm}

\usepackage{multirow}
\usepackage{booktabs}
\usepackage{graphicx} 
\usepackage{subcaption}
\usepackage{wrapfig}
\usepackage{makecell}
\usepackage{enumitem}
\usepackage{listings}
\usepackage{xcolor}
\usepackage{tabularx}
\usepackage{array}
\usepackage{adjustbox}
\usepackage{caption}

\theoremstyle{definition}

\theoremstyle{plain}

\newtcolorbox{keymessage}[1][Key message]{
  enhanced, breakable,
  colback=CosmosVioletPale, colframe=CosmosViolet, boxrule=0.6pt,
  borderline west={2.2pt}{0pt}{CosmosViolet},
  arc=1mm, title={#1}, fonttitle=\sffamily\bfseries, coltitle=white,
  left=4mm, right=4mm, top=2mm, bottom=2mm,
  before skip=0.6em, after skip=0.6em
}
\newtcolorbox{methodnote}[1][Method note]{
  enhanced, breakable,
  colback=CosmosVioletPale, colframe=CosmosViolet!45, boxrule=0.5pt,
  colbacktitle=CosmosVioletPale, coltitle=CosmosViolet,
  arc=1mm, title={#1}, fonttitle=\sffamily\bfseries,
  left=4mm, right=4mm, top=2mm, bottom=2mm,
  before skip=0.6em, after skip=0.6em
}

\newlength{\WideExtension}
\title{AI Cosmos: An Agentic Infra for Situated Learning}

\author{AI Cosmos Team}

\affiliation[]{Tsinghua University \\
(For the complete list of authors, please refer to the Contribution section)}

\abstract{

Large language models (LLMs) can improve task performance during learning, yet these gains do not necessarily reflect learners' ability to thereafter apply the knowledge and skills independently. We present AI Cosmos (\url{https://aicosmos.ai/}), an agentic infrastructure for situated learning that organizes learning around two complementary principles: scenario-based participation and cumulative assets of practice. AI Cosmos organizes learners and role-specialized agents within learning scenarios that integrate task-relevant knowledge and, when needed, executable workbenches. An asset–memory loop preserves the processes and outcomes of practice as reusable assets with explicit provenance and dependencies. We evaluate scenario-based interaction through two preregistered randomized controlled studies examining the acquisition of legal reasoning and programming competence, holding the same underlying LLM across conditions. AI Cosmos achieves greater learning gains than a conventional LLM in legal reasoning and higher mastery of unseen programming problems than an LLM-as-Teacher baseline.

}

\date{\today}
\correspondence{\email{zjf23@mails.tsinghua.edu.cn}}
\metadata[Code]{\url{https://github.com/THU-AICosmos}}

\begin{document}

\maketitle


\begin{flushright}
\itshape
``Education becomes transformative when knowledge is not merely expressed,\\
but enacted, shared, and turned into reusable intellectual assets.'' \\
-- Manifesto of AI Cosmos
\end{flushright}


\begin{figure*}[htbp]
    \centering
    \includegraphics[width=.95\linewidth]{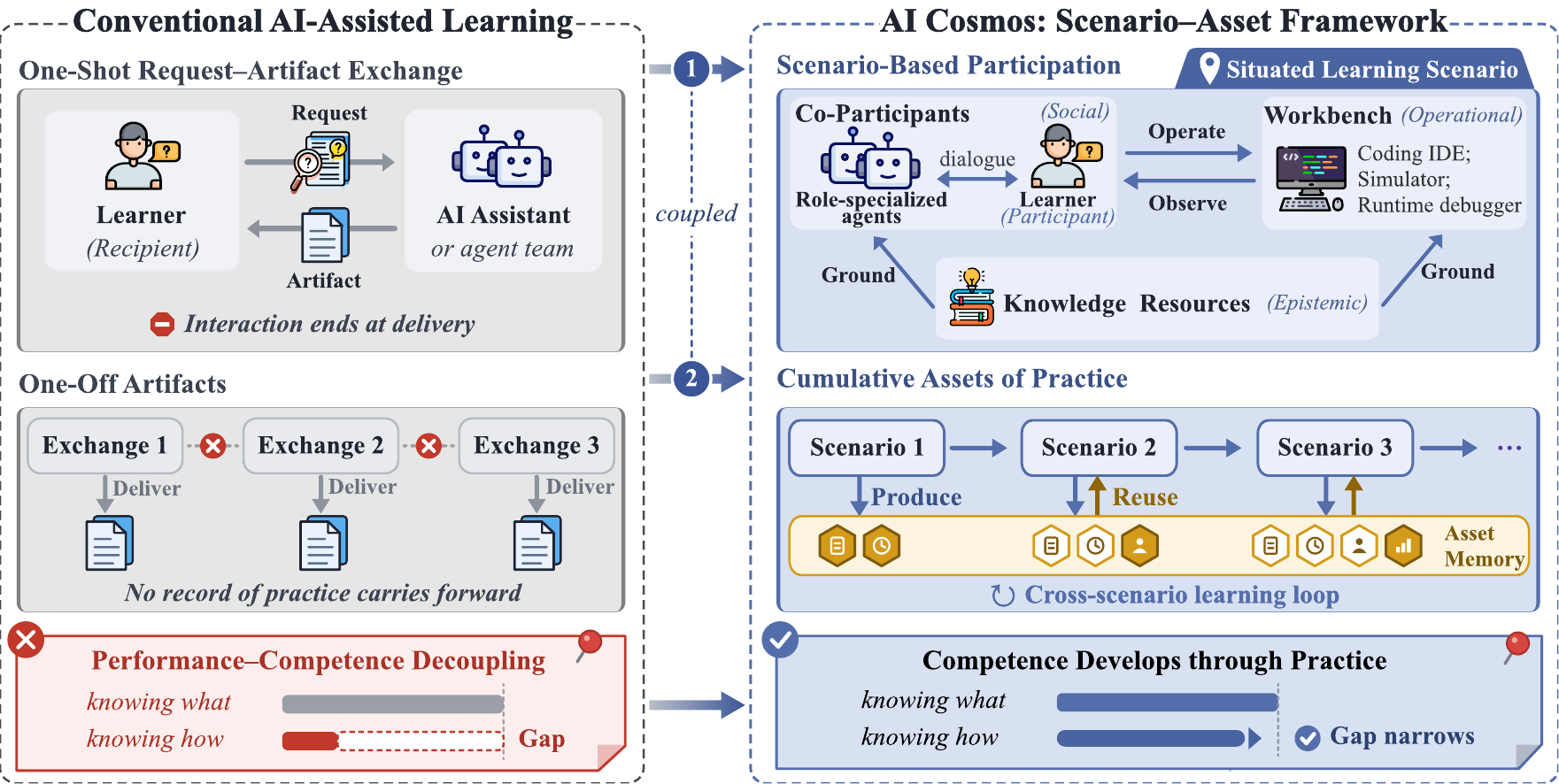}
    \caption{Overview of the Scenario–Asset Framework. AI Cosmos shifts AI-assisted learning from request–artifact exchange to scenario-based participation, and from one-off outputs to cumulative assets of practice, supporting competence development through sustained practice.}
    \label{fig:intro_paradigm_comparison}
\end{figure*}

\section{Introduction}

In 1929, philosopher Alfred North Whitehead described the problem of \textbf{inert knowledge} in his seminal collection ``The Aims of Education'': 
\textit{``...the problem of keeping knowledge alive, of preventing it from becoming inert, which is the central problem of all education''} \citep{whitehead1929aims}. 
Nearly a century later, Whitehead's diagnosis remains relevant, despite successive advances in educational technologies.  
Today, generative AI appears to offer a promising remedy to this enduring problem by providing intelligence that is continuously accessible, highly interactive, and readily adaptable into explanations and personalized tutorials \citep{ruiz2023empowering,baidoo2023education,ooi2025potential,characterglm}.

Yet, this remedy may not close the gap between \textbf{\emph{knowing}} and \textbf{\emph{doing}} but instead widen it in a new form.
Across educational settings, learners become more fluent in reproducing AI-provided concepts, explanations, and solutions \citep{anthropic_education} without commensurate gains in their ability to apply that knowledge independently \citep{metacognitive_laziness,generative_ai_harm_learning}, a discrepancy we refer to as \textbf{performance--competence decoupling} \citep{performance_competence_learning}.
For instance, an undergraduate student may be able to explain the divide-and-conquer tactic of quicksort yet struggle to implement it robustly \citep{llm_harm_learning_in_code}; similarly, a bar-exam candidate can fluently state a legal doctrine yet fails to apply it to an unfamiliar case \citep{legal_education}. 
These examples reflect a structural asymmetry: \textit{generative AI has substantially reduced the cost of accessing and reproducing declarative knowledge (``knowing what'') \citep{know_what_cheaper}, without producing an equivalent improvement in procedural knowledge (``knowing how'') \citep{ai_reduce_cognitive_effort}.}

We should note that this asymmetry no longer stems from LLM capability, but from the interaction scheme in which these models are used \citep{humam_ai_interaction_design,human_ai_interaction_structure}. 
Despite the diversity of AI-assisted educational applications, from content creation to lesson planning \citep{maic,lesson_planner,ppt_generation}, a common interaction pattern is the \textbf{request--artifact exchange}, where the learner submits a request, the system produces an explanation or artifact, and the interaction often ends once that artifact is delivered \citep{epistemic_outsourcing}.
Question answering is such an epistemic outsourcing example, more accurate and pedagogically appropriate responses may improve immediate task performance without requiring learners to reason through, test, or revise solutions themselves.
The limitation, therefore, is not simply the quality of generated outputs, but whether the interaction supports the activities through which competence develops. 
Learning research emphasizes active participation, feedback, and iterative revision within meaningful, scaffolded activities \citep{icap_framework,situated_learning,cognitive_apprenticeship,problem_based_learning}. 
When assistance is organized primarily around output delivery, learners' actions, feedback, and revisions may remain outside the interaction, limiting opportunities to support continued practice and use its outcomes to inform subsequent learning \citep{university_teaching,dialogic_pedagogy}.

Recent LLM-based tutoring systems structure pedagogical dialogue and feedback to support learners' reasoning \citep{autotutor,socratic_playground}, while general-purpose multi-agent frameworks coordinate role-specialized agents to complete complex tasks \citep{autogen,metagpt,chatdev}. 
Yet the effectiveness of these systems in supporting learners' sustained competence development cannot be inferred from the quality of instructional guidance or the success of agent task execution alone.
One consideration is \textbf{agency-preserving participation}, which emphasizes learners' active engagement in explaining, testing, and revising their own work within authentic or scaffolded activities, rather than merely receiving generated solutions \citep{situated_learning,icap_framework}.
Tool-using agents are often designed for autonomous task execution \citep{react,swe_agent}; when applied to educational settings, such designs may position learners primarily as requesters or evaluators of completed work.
A second consideration is maintaining \textbf{\textit{cumulative, inspectable record of practice}}. Constructionism emphasizes the educational value of artifacts that learners can create, examine, and improve over time \citep{constructionism}. 
Agentic systems also accumulate and refine reusable skills from past experience \citep{skillevolver}, although their reported gains primarily concern subsequent agent task performance rather than learners' use of prior practice. 
Taken together, these considerations motivate an approach to AI-assisted learning that centers learners' own reasoning and action within learning activities, while enabling subsequent activities to build on prior practice.

We translate these considerations into two complementary architectural shifts.
The first shift moves from \textit{request-artifact exchange} to \textit{scenario-based participation}, which organizes learning around learners' active involvement in socially mediated, tool-supported activities \citep{situated_learning,activity_theory}.
Each scenario integrates three configurable dimensions of practice: \textbf{social} (learners and role-specialized agents as co-participants), \textbf{epistemic} (task-relevant knowledge resources), and \textbf{operational} (stateful, executable workbenches, i.g., a coding IDE,, when needed).
Learners and agents coordinate through dialogue and, when applicable, perform task-relevant operations on shared workbench states.
The resulting observations inform subsequent dialogue and actions, supporting learners in applying knowledge, evaluating outcomes, and adapting their approaches through practice.
The second shift moves from \textit{one-off artifacts} to \textit{cumulative assets of practice}. 
Rather than treating final outputs as the endpoint of learning activities, this shift makes the processes and outcomes of practice available for reuse. 
Subsequent scenarios can thus build on accumulated experience, supporting continuity across learning activities.
Together, these shifts place learners' active participation and the accumulation of practical experience at the center of AI-assisted learning, shifting the emphasis from task completion to the development of learners' competence \citep{experiential_learning_cycle,reflective_practitioner}.

We instantiate these shifts in AI Cosmos (\url{https://aicosmos.ai/}), an agentic infrastructure for situated learning that organizes learner–agent participation in structured scenarios and preserves records of practice for reuse. 
A scenario control plane configures each learning activity from a learner's request by specifying its objective, participant roles, relevant knowledge resources, and executable workbenches when needed.
A task execution plane coordinates dialogue and workbench operations over shared task states, dynamically scheduling learners and agents to participate in iterative problem solving, execution, and revision.
To connect successive scenarios, an asset–memory loop extracts and validates reusable assets from scenario configurations and records of practice. 
By retaining provenance and dependencies, these assets make prior outcomes and the associated reasoning, feedback, and revisions available for inspection and adaptation in subsequent activities.
We evaluate scenario-based interaction through two preregistered randomized controlled studies examining the acquisition of legal reasoning and programming competence, finding greater learning gains than a conventional LLM in the former and stronger programming mastery than an LLM-as-Teacher baseline in the latter.

\vspace{-1mm}

\section{The Scenario–Asset Framework}

\subsection{The Scenario}

\textbf{Definition 2.1 (Scenario).} \textit{A scenario is a configured learning activity specified by the tuple $\mathcal{S} = \langle o_{\mathcal{S}}, \mathcal{P}_{\mathcal{S}}, \mathcal{K}_{\mathcal{S}}, \mathcal{W}_{\mathcal{S}} \rangle$, where $o_{\mathcal{S}}$ is the learning objective around which the activity is organized, and the remaining components operationalize three complementary dimensions of practice:}
\begin{itemize}[leftmargin=*, topsep=0pt, itemsep=0pt, partopsep=-5pt, parsep=0pt]
\item \textit{$\mathcal{P}_{\mathcal{S}}$ (\textbf{social}): The participant configuration, comprising role-specialized agents $\mathcal{P}^{\mathrm{agent}}_{\mathcal{S}}$ (e.g., explainer, mentor) and human learners $\mathcal{P}^{\mathrm{human}}_{\mathcal{S}}$. Their assigned roles determine how responsibility and action are distributed within the scenario;}
\item \textit{$\mathcal{K}_{\mathcal{S}}$ (\textbf{epistemic}): The task-relevant knowledge resources available to participants are queryable, e.g., reference materials, datasets, and retrieved from memory;}
\item \textit{$\mathcal{W}_{\mathcal{S}}$ (\textbf{operational}): The set of stateful workbenches through which participants act, e.g., a coding IDE. Each $w\in\mathcal{W}_{\mathcal{S}}$ maintains a mutable state and exposes operations for inspecting, creating, modifying, executing, or evaluating task-relevant objects.}
\end{itemize}

The epistemic and operational components may be empty, where $\mathcal{W}_{\mathcal{S}}=\varnothing$, participation is entirely discursive, and $\mathcal{K}_{\mathcal{S}}=\varnothing$, no external knowledge resource is explicitly bound to the activity.
A scenario proceeds through three stages: \textit{initialization} configures its components, \textit{execution} advances the scenario through successive state transitions directed toward $o_{\mathcal{S}}$, and \textit{completion} packages its residue and outputs for asset extraction.

\subsection{The Asset}

\textbf{Definition 2.2 (Asset).} \textit{An asset is a typed, provenance-bearing unit of reusable outcome, $[a=\langle \tau_a,c_a,s_a,\mathcal{A}_{a}^{\mathrm{prior}}\rangle]$, where $\tau_a$ is its type, $c_a$ is its content, $s_a$ identifies its originating scenario, and $\mathcal{A}_{a}^{\mathrm{prior}}$ contains the prior assets on which it directly depends. We distinguish five asset types:}
\begin{itemize}[leftmargin=*, topsep=0pt, itemsep=0pt, partopsep=-5pt, parsep=0pt]
\item \textit{\textbf{Role specification}: a reusable definition of a participant role, including its responsibilities, behavioral instructions, and applicable task conditions;}
\item \textit{\textbf{Scenario specification}: a complete or partial configuration $\langle o,\mathcal{P},\mathcal{K},\mathcal{W}\rangle$ that can be instantiated or adapted in subsequent scenarios;}
\item \textit{\textbf{Knowledge artifact}: a task-relevant object created or revised through a workbench, such as a slide deck, annotated document, or code module;}
\item \textit{\textbf{Process record}: an inspectable sequence of participant actions, feedback, critiques, and revisions through which an outcome emerged;}
\item \textit{\textbf{Experience trace}: structured evidence abstracted from execution, e.g., operation outcomes, failure cases, or observed relationships between configuration choices and task performance.}
\end{itemize}

By preserving content together with its provenance and dependencies, assets can be inspected, retrieved, adapted, and extended across scenarios. 
In particular, scenario specifications, process records, and experience traces retain the conditions and actions surrounding an outcome, making records of prior practice available to future participants rather than preserving only the final product.

\subsection{The Asset–Memory Feedback Loop}
\label{sec:feedback_loop}

\textbf{Definition 2.3 (Asset Memory).} \textit{At system update $t$, the asset memory $\mathcal{M}_t$ is the set of admitted assets, denoted as a asset-dependency graph $\mathcal{G}_t=(\mathcal{M}_t,\mathcal{E}_t)$, where $\mathcal{E}_t=\{(a',a)\mid a\in\mathcal{M}_t; a'\in\mathcal{A}_{a}^{\mathrm{prior}}\}.$ An edge $a'\to a$ denotes that $a$ depends on or is derived from $a'$. As dependencies refer only to prior assets, $\mathcal{G}_t$ forms a directed acyclic graph.}
Scenarios interact with asset memory via three operations:
\begin{itemize}[leftmargin=*, topsep=0pt, itemsep=0pt, partopsep=-5pt, parsep=0pt]
\item \textbf{Retrieval}: At initialization and, when needed, during execution $R(q_{\mathcal{S},t},\mathcal{M}_t) \to~\mathcal{A}^{\mathrm{in}}_{\mathcal{S},t} \subseteq \mathcal{M}_t$.
The query is initialized from the learning objective and scenario configuration and may subsequently incorporate the evolving  scenario's runtime state. Retrieved assets can configure roles, supply knowledge resources, or inform the selection and use of workbenches.

\item \textbf{Production}: During execution, the system records artifacts, participant interactions, and workbench traces. At completion, the system distills these records into candidate assets, $\mathcal{S} \to~\mathcal{A}^{\mathrm{out}}_{\mathcal{S}}$, each of which records its originating scenario and dependencies on retrieved assets or earlier candidates.

\item \textbf{Admission}: 
Candidates are evaluated using a type-specific predicate $\mathcal{V}_{\tau_a}(a)\in\{0,1\}$. Let $\mathcal{A}^{+}_{\mathcal{S}}$ be the largest dependency-closed subset of validated candidates: $\mathcal{A}^{+}_{\mathcal{S}}=\{a\in\mathcal{A}^{\mathrm{out}}_{\mathcal{S}} \mid \mathcal{V}_{\tau_a}(a)=1\}$.
The memory is updated as $\mathcal{M}_{t+1}=\mathcal{M}_t\cup\mathcal{A}^{+}_{\mathcal{S}}$, with the related dependency edges added to $\mathcal{E}_{t+1}$.

\end{itemize}

With these three operations, later scenarios can inspect, compose, and extend validated contributions from prior activities, enabling assets and records of practice to accumulate across interactions.
\section{System Architecture of AI Cosmos}
\label{sec:architecture}

\begin{figure*}[t]
    \centering
    \includegraphics[width=\linewidth]{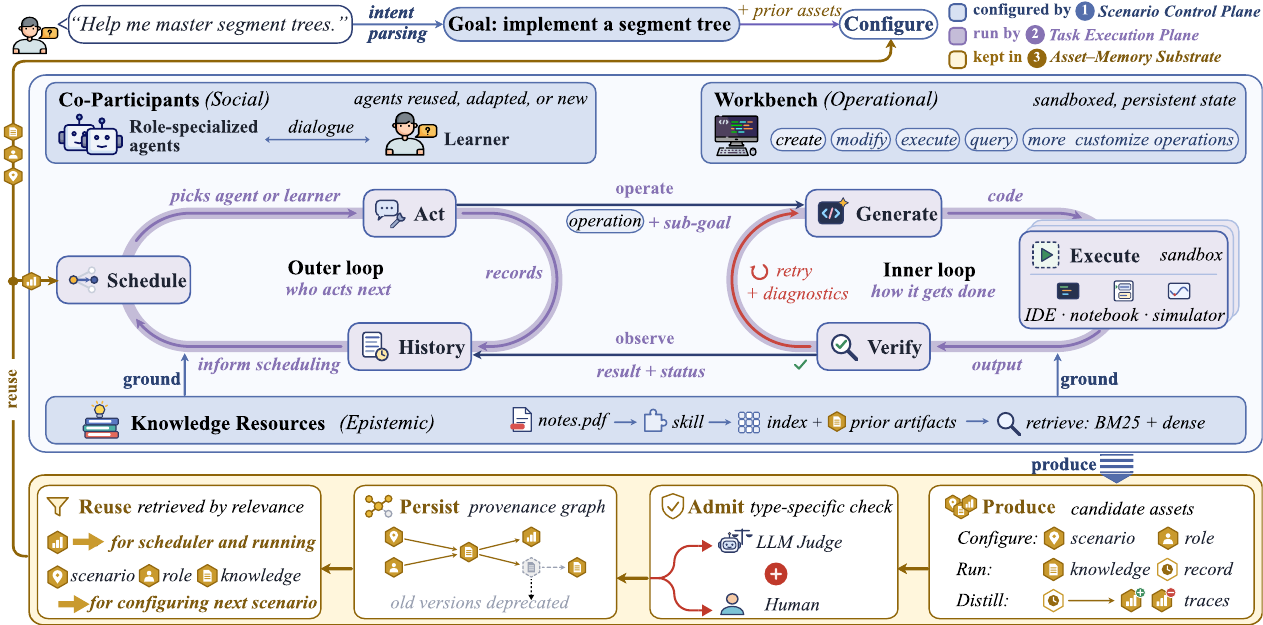}
    \caption{ Architecture of AI Cosmos. The scenario control plane configures learning environments, the task execution plane coordinates participant interactions and workbench operations, and the asset–memory substrate preserves and reuses validated assets across scenarios.}
    \label{fig:method}
\end{figure*}

AI Cosmos implements the Scenario--Asset Framework via three interconnected components (Figure \ref{fig:method}). 
The \textit{scenario control plane} configures learning objectives, participant roles, knowledge resources, and optional workbenches from learner requests and relevant prior assets. 
The \textit{task execution plane} coordinates learner--agent interactions and workbench operations via history-dependent scheduling.
Learners and agents operate on shared workbench states, with resulting observations informing subsequent interactions. 
The \textit{asset--memory substrate} derives reusable assets from scenario configurations and records of practice, manages their validation and admission, and preserves provenance and dependencies for reuse in subsequent scenarios.
 
\subsection{The Scenario Control Plane}
\label{sec:scene_control_plane}
 
This plane instantiates the scenario-configuration operator $\mathcal{F}_{\text{config}}$, mapping an initial learning request to an executable scenario for the task execution plane. 
Rather than relying on a fixed agent team and tool set, it configures the three dimensions according to the learning objective.
Let $u$ be a user input, comprising a request and a (possibly empty) set $\mathcal{U}$ of uploaded resources. The operator is defined as:
\begin{equation}
\mathcal{F}_{\text{config}}(u\,;\, \mathcal{M}) \;\longmapsto\; \mathcal{S} = \langle o_{\mathcal{S}}, \mathcal{P}_{\mathcal{S}}, \mathcal{K}_{\mathcal{S}}, \mathcal{W}_{\mathcal{S}} \rangle .
\end{equation}
It first grounds the request into the objective $o_{\mathcal{S}}$ through intent parsing.
This objective then conditions the scenario configuration that may reuse previously admitted assets through relevance-gated retrieval:
$\mathcal{A}^{\text{in}}_{\mathcal{S}} = R(o_{\mathcal{S}}, \mathcal{M})$, which returns only assets whose relevance to $o_{\mathcal{S}}$ exceeds a predefined threshold, otherwise $\mathcal{A}^{\text{in}}_{\mathcal{S}}=\varnothing$.
We write $\mathcal{A}^{\text{in},\mathcal{P}}_{\mathcal{S}}$, $\mathcal{A}^{\text{in},\mathcal{K}}_{\mathcal{S}}$, and $\mathcal{A}^{\text{in},\mathcal{W}}_{\mathcal{S}}$ for the asset subsets used in three dimensions.

\textbf{Social configuration.}
This dimension is realized as a shared, multiparty interaction space comprising both agents and human participants: $\mathcal{P}_{\mathcal{S}} = \mathcal{P}^{\text{agent}}_{\mathcal{S}} \cup \mathcal{P}^{\text{human}}_{\mathcal{S}}$.
Given the objective $o_{\mathcal{S}}$ and any retrieved role specifications, it determines the roles required by the scenario and instantiates one agent per role, with both the number of roles $n$ and each profile determined adaptively:
\begin{equation}
    \{r_1, \dots, r_n\} = \mathrm{Roles}\!\left(o_{\mathcal{S}},\, \mathcal{A}^{\text{in},\mathcal{P}}_{\mathcal{S}}\right) ,
\quad
\mathcal{P}^{\text{agent}}_{\mathcal{S}} = \{\, \mathrm{agent}(r_i) \,\}_{i=1}^{n} .
\end{equation}
Each role profile $r_i$ specifies the participant's task responsibility, expertise, interaction policy, and access to knowledge resources and workbench operations.
When $\mathcal{A}^{\text{in},\mathcal{P}}_{\mathcal{S}}\neq\varnothing$, $\mathrm{Roles}(\cdot)$ reuses or adapts roles from previously admitted configurations; otherwise, it synthesizes new role profiles from scratch.

\textbf{Epistemic configuration.}
This dimension is realized as a knowledge base that remains queryable throughout scenario execution. 
AI Cosmos holds a library $\Sigma$ of modular skills that can be loaded on demand \citep{claude_skill}.
For each uploaded resource $\rho \in \mathcal{U}$, the control plane selects a compatible skill $\sigma_\rho \in \Sigma$ and applies it to produce a set of indexed, retrievable units $\sigma_\rho(\rho)$.
Let $\mathcal{A}^{\text{in},\mathcal{K}}_{\mathcal{S}} \subseteq \mathcal{A}^{\text{in}}_{\mathcal{S}}$ denote the retrieved knowledge-bearing assets, and $\mathrm{Index}(\mathcal{A}^{\mathrm{in},\mathcal{K}}_{\mathcal{S}})$ denote the indexed units derived from the retrieved assets.
The knowledge base is 
\begin{equation}
\mathcal{K}_{\mathcal{S}}
=
\underbrace{
\bigcup_{\rho \in \mathcal{U}}
\sigma_{\rho}(\rho)
}_{\text{units derived from uploaded resources}}
\;\cup\;
\underbrace{
\mathrm{Index}\!\left(
\mathcal{A}^{\mathrm{in},\mathcal{K}}_{\mathcal{S}}
\right)
}_{\text{units inherited from prior scenarios}}.
\end{equation}
During execution, authorized participants access $\mathcal{K}_{\mathcal{S}}$ through a retrieval service. When neither uploaded resources nor previously accumulated knowledge is required, $\mathcal{K}_{\mathcal{S}} = \varnothing$.

\textbf{Operational configuration.}
This dimension is realized by selecting and binding workbenches through which participants execute task-relevant objects.
Let \(\mathcal{W}\) denote the registry of available workbenches (e.g., a coding IDE or domain-specific instrument), the plane selects the subset required by the objective:
\begin{equation}
\mathcal{W}_{\mathcal{S}}
=
\mathrm{Select}\!\left(
o_{\mathcal{S}},
\mathcal{W},
\mathcal{A}^{\mathrm{in},\mathcal{W}}_{\mathcal{S}}
\right)
\subseteq
\mathcal{W}.
\end{equation}
For a purely discursive objective, $\mathcal{W}_{\mathcal{S}} = \varnothing$.
Each workbench $w \in \mathcal{W}_{\mathcal{S}}$ maintains a shared, mutable state $x_w$ and exposes a typed action interface $\mathcal{I}_w$.
The interface consists of named operations for inspecting or transforming that state, such as creating, modifying, executing, or querying a task-relevant object.
Binding a workbench connects its action interface to the scenario's shared interaction stream.
An authorized agent invokes an operation in $\mathcal{I}_w$ through a structured tool call, whereas a human participant acts through a graphical interface whose interactions are mapped to the same underlying operations.
An operation updates $x_w$ and emits an observation that is recorded in the shared interaction history.
Thus, conversation and operation remain synchronized: dialogue can initiate changes to the workbench, and workbench observations can immediately inform subsequent dialogue and action.
The access scope encoded in each role profile $r_i$ determines which workbenches in $\mathcal{W}_{\mathcal{S}}$, and which operations in their $\mathcal{I}_w$, the corresponding agent is authorized to invoke.

\subsection{The Task Execution Plane}
\label{sec:tep}

The task execution plane runs a configured scenario $\mathcal{S}$ toward its objective $o_{\mathcal{S}}$ by coupling deliberation in the shared interaction space with action in the bound workbenches.

\textbf{Outer loop: multi-party scheduling.}
At each outer step, the execution plane selects the next system-mediated participant according to the current interaction history $H_t$ and objective:
\begin{equation}
p_t = \mathrm{Schedule}(H_t,\, o_{\mathcal{S}}) \in \mathcal{P}_{\mathcal{S}}.
\end{equation}
The selected participant contributes to the shared interaction or invokes a bound workbench.
Since scheduling is conditioned on the evolving history rather than a fixed turn order, the scenario can move dynamically among explanation, questioning, critique, tool use, and human handoff.
All agent actions remain subject to the role and access policies established by the scenario control plane.
Each scheduled action and its resulting observation are recorded in the history and inform subsequent scheduling.

\textbf{Grounding agent turns with hybrid retrieval.}
When a knowledge base $\mathcal{K}_{\mathcal{S}}$ is available, the task execution plane derives a query $q_t^{\mathcal{K}}$ from the interaction history $H_t$ and scores each entry $k$ by combining normalized BM25 \citep{bm25} and embedding-based cosine similarity:
\begin{equation}
\mathrm{score}\!\left(q_t^{\mathcal{K}},k\right)
=
\lambda\,
\widetilde{\mathrm{BM25}}
\!\left(q_t^{\mathcal{K}},k\right)
+
(1-\lambda)\,
\widetilde{\mathrm{Cos}}
\!\left(
\mathbf{e}_{q_t^{\mathcal{K}}},
\mathbf{e}_k
\right),
\end{equation}
where $\lambda\in[0,1]$ balances lexical and semantic relevance, and both similarity scores are normalized to a common range. Entries with $\mathrm{score}(q_t^{\mathcal{K}},k)\geq\tau$ are provided as grounding context for subsequent agent actions.

\textbf{The workbench: sandboxed state and intent-level operations.}
Each workbench $w\in\mathcal{W}_{\mathcal{S}}$ comprises an isolated execution environment, which maintains its persistent state $x_w$ and a typed action interface $\mathcal{I}_w$.
The sandbox preserves task state across operations and confines untrusted model-generated execution from the host environment and other scenarios.
The interface exposes a small vocabulary of operations:
\begin{equation}
\mathcal{I}_w \subseteq \mathcal{I} = \{\,\textsc{create},\ \textsc{modify},\ \textsc{execute},\ \textsc{query},\ \dots\,\}.
\end{equation}
Here, \textsc{create} creates or initializes a task-relevant object, 
\textsc{modify} changes an existing object,
\textsc{execute} runs a computation or executable object, and
\textsc{query} inspects the current state.
Each workbench implements only the operations supported by its capabilities.
For agent invocations, operations are specified at the level of intent: the agent provides an operation and a natural-language sub-goal, which a workbench-specific adapter translates into platform primitives \citep{embodiedact}.
Thus, heterogeneous environments can share the same higher-level scheduling and execution logic while differing only in their runtime and adapter.

\textbf{Inner loop: executing a workbench operation.}
When a scheduled agent action requires workbench $w$, the agent invokes an operation $\iota_w \in \mathcal{I}_w$, authorized by its role profile $r_i$, with a sub-goal $o_g$ through a structured tool call \citep{react}. The operation executes within the sandbox through iterative generation and execution of platform-specific primitives, using diagnostic feedback to guide retries. Let $j$ index execution attempts, with $x_w^{(0)}=x_w$ denoting the initial workbench state. At each attempt,
\begin{equation}
c_w^{(j)}
=
\mathrm{Gen}_w
\!\left(
\iota_w,o_g,x_w^{(j)},\epsilon^{(j)}
\right),
\qquad
\left(
x_w^{(j+1)},y_w^{(j)}
\right)
=
\mathrm{Exec}_w
\!\left(
x_w^{(j)},c_w^{(j)}
\right),
\end{equation}
where $\mathrm{Gen}_w$ is an LLM-based generator of workbench-specific primitives $c_w^{(j)}$ (e.g., Python cells in Jupyter), while $\mathrm{Exec}_w$ executes these primitives in an isolated sandbox, yielding the updated state $x_w^{(j+1)}$ and observation $y_w^{(j)}$. 
The diagnostic input $\epsilon^{(j)}$ captures feedback from the preceding failed attempt, with $\epsilon^{(0)}=\varnothing$. After each attempt, the workbench evaluates the result against the sub-goal:
\begin{equation}
\mathrm{Verify}\!\left(o_g,y_w^{(j)}\right)\in\{0,1\}.
\end{equation}
If verification fails and the retry budget permits, diagnostic feedback guides the next generation attempt. 
The loop terminates upon successful verification or budget exhaustion. 
The complete invocation trace, including intermediate failures and repairs, is retained for later inspection and asset extraction, while the terminal observation and execution status are incorporated into the shared interaction history $H_{t+1}$ without exposing every intermediate attempt.

\textbf{Interruptible human participation.}
Human participants may intervene during scenario execution without waiting for scheduler-selected turns. 
Subject to scenario permissions, they can inspect or modify bound workbenches through a graphical interface, operating on the same persistent state and underlying operations as authorized agents. 
These interventions are recorded in the shared interaction history, allowing subsequent scheduling and agent actions to account for the updated workbench state.

\subsection{The Asset–Memory Substrate}
\label{sec:substrate}

The two planes above orchestrate an individual scenario, while the asset--memory substrate connects successive scenarios through the feedback loop in \S\ref{sec:feedback_loop}, which manages four stages of the asset lifecycle.

\textbf{Production: deriving assets from practice.}
At \emph{configuration}, each instantiated role profile $r_i$ yields a
\emph{role specification}, and the configured tuple $\langle o_{\mathcal{S}}, \mathcal{P}_{\mathcal{S}}, \mathcal{K}_{\mathcal{S}}, \mathcal{W}_{\mathcal{S}} \rangle$ yields a \emph{scenario specification}.
During \emph{execution}, durable task-relevant objects created or materially revised in a workbench become candidate \emph{knowledge artifacts}, while dialogue, retrieval events, workbench operations, observations, failures and repairs, and human interventions accumulate in the execution history. 
At completion, the resulting history $H_{\mathcal{S}}$ is persisted as a \emph{process record}.
An offline pass further distills reusable \emph{experience traces}:
\begin{equation}
\mathrm{Distill}(H_{\mathcal{S}})
=
\left\{
\langle \kappa_\eta,\beta_\eta,\ell_\eta\rangle
:
\eta\in\mathrm{Seg}(H_{\mathcal{S}})
\right\}.
\end{equation}
Here, $\mathrm{Seg}(\cdot)$ partitions the execution record at dialogue-turn and workbench-attempt boundaries.
For each episode $\eta$, $\kappa_\eta$ summarizes the triggering context, $\beta_\eta$ abstracts the action taken, and $\ell_\eta\in\{+,-\}$ denotes its task-level outcome:
\begin{equation}
\ell_\eta =
\begin{cases}
\mathbb{1}^{\pm}
\!\left[
\mathrm{Prog}
\!\left(
H_{\mathrm{end}(\eta)},o_{\mathcal{S}}
\right)
>
\mathrm{Prog}
\!\left(
H_{\mathrm{start}(\eta)},o_{\mathcal{S}}
\right)
\right],
&
\eta \text{ is a dialogue turn},
\\[5pt]
\mathbb{1}^{\pm}
\!\left[
\mathrm{Verify}(o_g,y_w^{(j)})=1
\right],
&
\eta \text{ is a workbench attempt},
\end{cases}
\end{equation}
where $\mathrm{Prog}(\cdot)$ is the scheduler's progress assessment, $\mathrm{Verify}(\cdot)$ the inner-loop self-check, and $\mathbb{1}^{\pm}[\Delta]$ returns $+$ if $\Delta$ holds and $-$ otherwise.
A failed attempt often followed by a successful repair therefore yields distinct negative and positive traces.
These labels capture task-level utility rather than pedagogical effectiveness.

\textbf{Admission: type-specific validation.}
Process records are admitted without quality filtering to preserve execution histories. For other candidate assets, an LLM-based evaluator applies a type-specific acceptance rubric $\Phi_{\tau_a}(o_{\mathcal{S}})$, with access to asset content, provenance, and scenario context:
\begin{equation}
\mathcal{V}_{\mathrm{auto}}(a)
=
\bigwedge_{\phi\in\Phi_{\tau_a}(o_{\mathcal{S}})}
\mathrm{Judge}(\phi,a),
\qquad
\mathrm{Judge}(\phi,a)\in\{0,1\},
\end{equation}
where $\tau_a$ denotes the asset type. Knowledge artifacts additionally require human approval. The final admission criterion is
\begin{equation}
\mathcal{V}(a)=
\begin{cases}
1, & \text{if }a\text{ is a process record},\\
\mathcal{V}_{\mathrm{auto}}(a)\land\mathrm{HumanGate}(a),
& \text{if }a\text{ is a knowledge artifact},\\
\mathcal{V}_{\mathrm{auto}}(a), & \text{otherwise}.
\end{cases}
\end{equation}
Only assets satisfying $\mathcal{V}(a)=1$ are admitted to reusable memory.

\textbf{Persistence: memory and provenance.}
Each admitted asset is stored as $a=\langle \tau_a,c_a,s_a,\mathcal{A}^{\mathrm{prior}}_a,\mathrm{st}_a\rangle$, where $\mathrm{st}_a\in\{\texttt{active},\texttt{deprecated}\}$ denotes its lifecycle status. Assets form a directed provenance graph $\mathcal{G}$, with an edge $a'\rightarrow a$ for each dependency $a'\in\mathcal{A}^{\mathrm{prior}}_a$. Asset content and dependencies remain immutable: when a new asset $a''$ supersedes $a$, it references $a$ as a dependency, and $\mathrm{st}_a$ is set to \texttt{deprecated}. By default, retrieval is restricted to active assets:
\begin{equation}
\mathcal{M}_{\mathrm{active}}
=
\left\{
a\in\mathcal{M}
\mid
\mathrm{st}_a=\texttt{active}
\right\}.
\end{equation}
Deprecated assets remain available for provenance inspection and rollback. Since dependencies point from earlier to later assets, $\mathcal{G}$ remains acyclic.

\textbf{Reuse: asset-specific integration.}
During scenario configuration, retrieved scenario specifications can initialize new configurations, while role specifications can be reused or adapted. Validated knowledge artifacts are indexed into $\mathcal{K}_{\mathcal{S}}$ for retrieval, whereas process records remain available for provenance and inspection without being treated as trusted knowledge by default. During execution, experience traces are retrieved using a query $q_t^{\mathrm{exp}}$ derived from the current interaction history:
\begin{equation}
\mathcal{A}^{\mathrm{exp}}_t
=
R_{\mathrm{exp}}
\!\left(
q_t^{\mathrm{exp}},
\mathcal{M}_{\mathrm{active}}
\right).
\end{equation}
Positive traces identify actions that advanced similar task states, while negative traces denote actions associated with failure. 
These traces provide auxiliary context for $\mathrm{Schedule}(\cdot)$ without determining its decisions.

\textbf{Compounding via reuse.}
The substrate closes the feedback loop by allowing validated assets and execution experience from one scenario to inform subsequent scenarios.
When relevant prior assets exist, reuse can reduce repeated configuration and reconstruction while preserving the provenance of the reused material.

\section{Experiments}
\label{sec:experiments}
 
\subsection{Human Study: Situated Learning for Bar-Examination Preparation}
\label{sec:exp_legal}

Legal problem solving provides a demanding test of application-oriented learning: success requires not only recalling legal doctrine but also identifying relevant rules and applying them to novel fact patterns.
We therefore use bar-examination preparation to test whether restructuring LLM interaction from conventional chat into a structured scenario improves learning of applied legal reasoning.

\subsubsection{Study Design and Measures}

\textbf{Design and participants.}
We conducted a pre-registered, three-arm, parallel-group randomized controlled trial. 
A total of 131 students were assigned in a 1:1:1 ratio to one of three learning conditions:
(i) \textit{vanilla LLM}, a conventional request--response assistant, 
(ii) \textit{LLM-as-Teacher} in which the same LLM was prompted to adopt an expert-tutor,
or (iii) \textit{AI Cosmos}. 
All conditions used the same underlying LLM, holding model capability constant while varying the interaction architecture.
Following our preregistered data-integrity criteria, we excluded participants whose learning logs failed to meet minimum-engagement requirements (e.g., insufficient interaction or exceptionally short inputs), leaving an analyzed sample of $N=111$ ($37$ per arm). 
We observed no statistically significant between-group differences in the measured pre-intervention characteristics, including sex, prior legal training, age, AI literacy, self-efficacy, and pre-test performance (all $p\geq0.27$; Table \ref{legal_tab:barexam-baseline} in Appendix \ref{app_sec:exp_legal}).

\textbf{Procedure and measures.}
Participants completed four phases:
(1) \textit{Pre-test}: An AI-literacy inventory, the General Self-Efficacy Scale (GSES, \cite{gses}), and a legal-knowledge assessment
covering civil and criminal law. This assessment combined objective single- and multiple-choice items with open-response questions. 
(2) \textit{Learning}: Participants studied matched civil- and criminal-law problems using only their assigned system. 
(3) \textit{Immediate post-test}: A new, difficulty-matched legal assessment with the same composition, followed by the GSES and the NASA-TLX workload index \citep{nasatlx}. 
(4) \textit{Delayed post-test} ($\approx\!4$ weeks later): A randomly selected subsample ($n=18$ per condition) completed a new assessment of the same composition to probe retention. 
All legal assessments were scored on a 0--100 scale, with objective items weighted at $80\%$ and open-response items at $20\%$.
The open responses were scored by three law PhDs whose ratings showed high agreement (Cronbach's $\alpha=0.95$ pre-test, $0.96$ post-test) and were averaged. 
The outcome was the \emph{learning gain},  defined as the post-test minus pre-test change in total score.

\textbf{Statistical analysis.}
We report group means and standard deviations for all primary outcomes.
Unadjusted differences in learning gain were examined using one-way ANOVA\footnote{$F(\mathrm{df}_1,\mathrm{df}_2)$ denotes the numerator and denominator degrees of freedom, with $\mathrm{df}_1=k-1$ for $k$ groups and $\mathrm{df}_2=N-k$. 
Fitting the ANCOVA covariate consumes one further degree of freedom, giving $F(2,107)$. }, with the Kruskal--Wallis test as a nonparametric robustness check.
Shapiro--Wilk tests did not reject normality of the gain distributions (all $p>0.05$), and Levene's test indicated no evidence of unequal variances ($p=0.95$).
The primary between-group inference used ANCOVA on immediate post-test scores, with pre-test score entered as a covariate.
This approach adjusts for chance baseline variation and generally provides greater precision than analysis of raw change scores.
The homogeneity-of-regression-slopes assumption was supported by the condition-by-pre-test interaction test ($p=0.77$).
We report the omnibus ANCOVA effect together with $\eta^2$ and model-adjusted contrasts between AI Cosmos and the control conditions.
For complementary unadjusted comparisons of raw learning gains, we report Welch's $t$ tests and standardized effect sizes (Cohen's $d$ or
bias-corrected Hedges' $g$) with bootstrap $95\%$ confidence intervals \citep{cohen}.
Subgroup and assessment-component analyses were exploratory and were not adjusted for multiplicity.
All hypothesis tests were two-sided with $\alpha=0.05$.

\begin{figure*}[tbp]
    \centering
    \includegraphics[width=\linewidth]{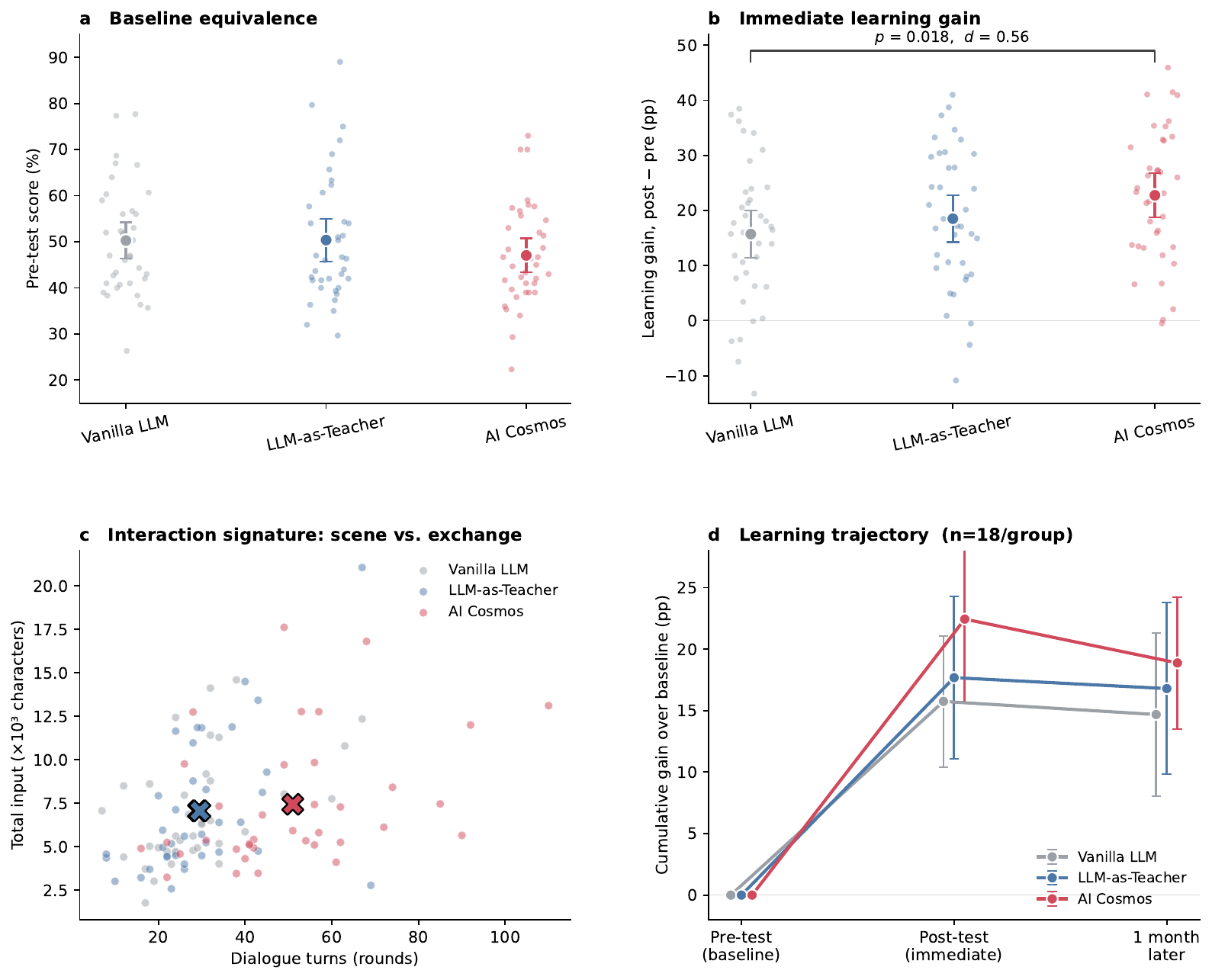}
    \caption{\textbf{Reorganizing the interaction as a structured scene improves bar-exam learning over a vanilla assistant.}
    Three arms ($n=37$ each) shared a single base model, differing only in interaction architecture.
    \textbf{(a) Pre-test scores.} Dots are individuals, markers denote mean\,$\pm$\,$95\%$\,CI. The arms were equivalent at baseline ($F(2,108)=0.88$, $p=0.42$), with AI Cosmos being numerically the lowest.
    \textbf{(b) Immediate learning gain (post\,$-$\,pre).} Gains increased monotonically across arms of increasing interaction structure ($15.7\!\to\!18.5\!\to\!22.8$\,pp; linear-trend contrast $p=0.018$). The bracket highlights the a priori primary contrast, AI Cosmos vs.\ vanilla ($\Delta=+7.0$\,pp, Cohen's $d=0.56$, bootstrap $95\%$\, CI $[0.10,1.05]$; $p=0.018$); an ANCOVA adjusting for the pre-test reproduced this finding ($+4.5$\,pp, $p=0.018$, partial $\eta^2=0.053$). The LLM-as-Teacher arm did not differ significantly from vanilla ($p=0.12$), locating the performance effect in the architecture rather than the persona prompt.
    \textbf{(c) Interaction dynamics.} Each point is one participant, and $\bm{\times}$ are group centroids. AI Cosmos elicited $\sim$$1.7\times$ more dialogue turns ($51.0$ vs.\ $29.2$/$29.6$; $F(2,108)=21.8$, $p<10^{-7}$, $\eta^2=0.29$, $g\!\approx\!1.2$) while requiring a statistically identical total input ($\sim$$7{,}100$ characters, $p=0.88$). This yielded $\sim 40\%$ fewer characters per turn ($162$ vs.\ $276$; $p\approx0.001$, $g=-0.80$), which is a behavioral signature of active participation replacing one-shot delivery.
    \textbf{(d) Cumulative gain} over each arm's own baseline at three time points within the delayed-test subsample ($n=18$ per arm; mean,$\pm$,$95\%$,CI). AI Cosmos retained the largest net gain at one month ($+18.9$\,pp). Note: this subsample analysis is exploratory, and the omnibus was not significant ($p=0.62$).
    }
    \label{legal_fig:overall}
    \vspace{-3mm}
\end{figure*}

\begin{figure*}[t]
\centering
\includegraphics[width=\linewidth]{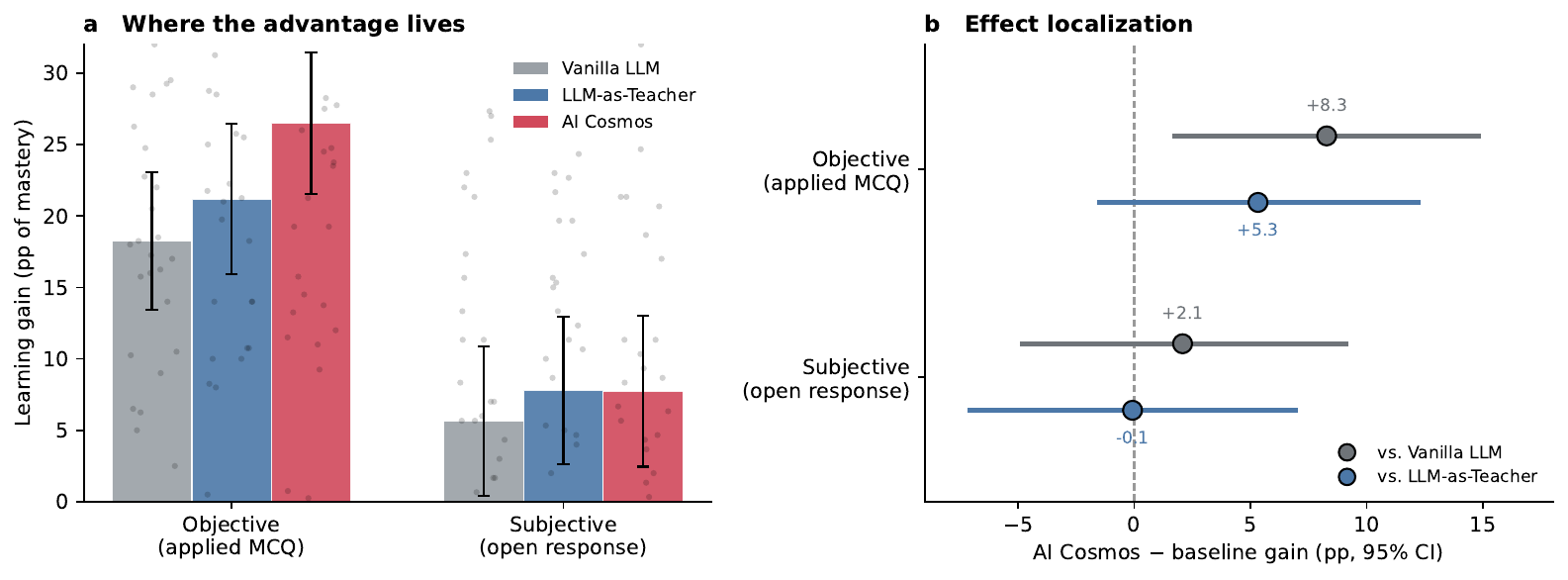}
    \caption{\textbf{The advantage concentrates in the applied scenario-based choice, rather than open-ended argumentation.}
    Each test was split into objective (scenario-based choice) and subjective (open-response) sections, each renormalized to $0$--$100$.
    \textbf{(a) Learning gain by arm and section} (bars indicate mean\,$\pm$\,$95\%$\,CI; points indicate individuals). On the objective section gains were ordered $18.2<21.2<26.5$\,pp ($F(2,108)=2.89$, $p=0.060$, $\eta^2=0.051$; ANCOVA-adjusted AI Cosmos vs.\ vanilla $p=0.038$). The subjective section improved equivalently across arms ($5.6$--$7.8$\,pp; $F(2,108)=0.23$, $p=0.80$).
    \textbf{(b) AI Cosmos\,$-$\,baseline gain for each section against both controls} (point estimate, $95\%$ bootstrap CI; grey indicates vs.\ vanilla; blue indicates vs.\ LLM-as-Teacher). On the objective section the advantage over vanilla excludes zero ($+8.3$\,pp $[1.8,14.9]$) and over the teacher baseline is positive but unresolved ($+5.3$\,pp $[-1.5,12.2]$). On the subjective section, neither contrast significantly differs from zero ($+2.1$ $[-4.9,9.3]$; $-0.1$ $[-7.1,7.1]$). The group\,$\times$\,section interaction in a mixed-design ANOVA was not significant ($F(2,108)=0.95$, $p=0.39$). Thus, the localization is descriptive. Inter-rater reliability for subjective scoring was high (Cronbach's $\alpha=0.95$--$0.96$).}
    \label{legal_fig:type}
    \vspace{-3mm}
\end{figure*}
\begin{figure*}[t]
\centering
    \includegraphics[width=\linewidth]{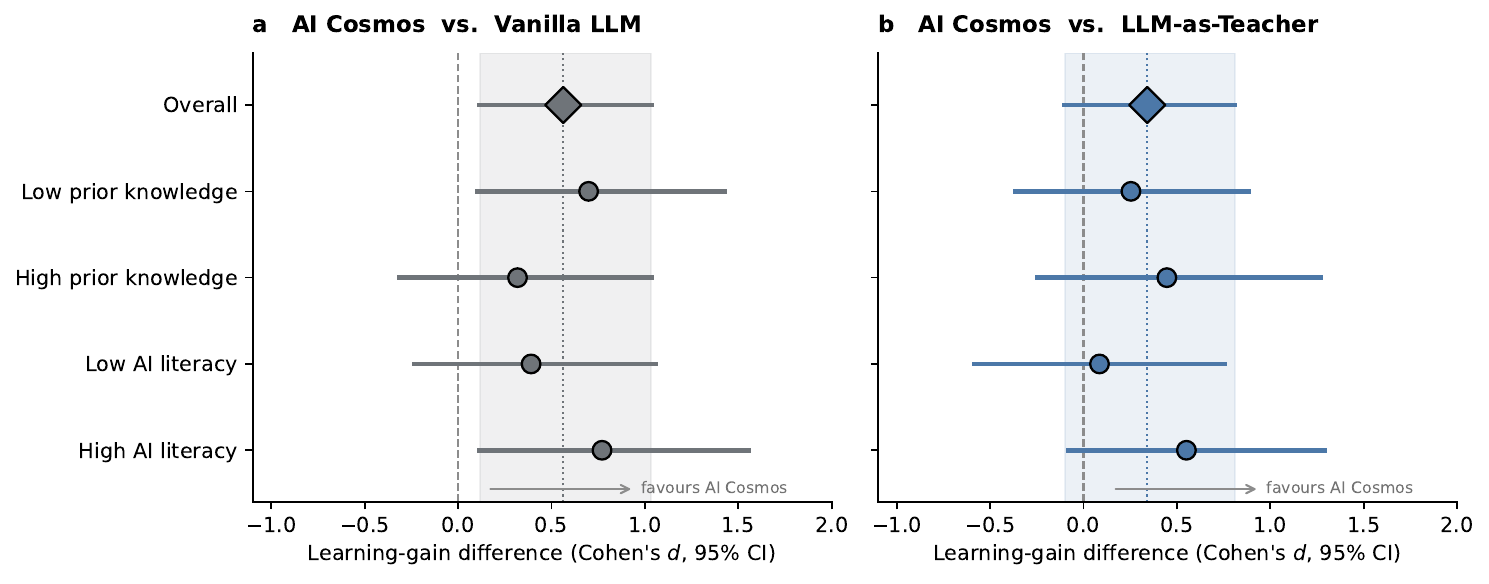}
    \caption{\textbf{The advantage over the vanilla assistant is directionally consistent across learner subgroups.}
    AI Cosmos\,$-$\,baseline difference in immediate learning gain (Cohen's $d$, $95\%$ bootstrap CI) against \textbf{a, the vanilla assistant} and \textbf{b, the LLM-as-Teacher}, for the full sample (diamond; shaded band and dotted line mark the overall estimate) and four strata from median splits on prior knowledge (median $=46.7$) and AI literacy. Against vanilla, the effect is positive in every stratum and most precisely estimated overall ($d=0.56$ $[0.10,1.05]$). Against the teacher baseline, it remains positive throughout but lacks statistical resolution at this sample size (overall $d=0.34$ $[-0.10,0.82]$). Subgroup cells contain $15$--$22$ participants per arm and are uncorrected for multiplicity. }
    \label{legal_fig:forest}
    \vspace{-3mm}
\end{figure*}

\subsubsection{Results}

\textbf{Learning efficacy.}
All three conditions showed large pre--post gains (paired $t$, all $p<10^{-8}$; $d_z=1.22$, $1.45$, and $1.88$ for the vanilla, LLM-as-teacher, and AI Cosmos). 
AI Cosmos yielded the largest raw gain ($22.8\pm12.1$\,pp), followed by LLM-as-teacher ($18.5\pm12.7$) and vanilla LLM ($15.7\pm12.9$; Figure~\ref{legal_fig:overall}b). 
The omnibus ANOVA on raw gains approached, but did not reach, statistical significance ($F(2,108)=2.94$, $p=0.057$, $\eta^2=0.052$).
However, the primary planned contrast was clear: 
AI Cosmos outperformed the vanilla LLM by $7.0$\,pp (Welch $t(71.8)=2.42$, $p=0.018$), a medium effect size ($d=0.56$, bootstrap $95\%$ CI $[0.10,1.05]$). 
An ANCOVA adjusting for the AI Cosmos group's slightly lower baseline reproduced this advantage on adjusted post-test means (i.e., post-test scores predicted at the common pre-test mean, i.e., adjusted means $70.2$ vs.\ $65.8$; group effect $F(2,107)=3.00$, $p=0.054$, partial $\eta^2=0.053$; AI Cosmos vs.\ vanilla $b=4.47$, $p=0.018$), with the pre-test serving as a significant covariate ($b=0.20$, $p=0.001$).
The advantage of AI Cosmos over the stronger LLM-as-teacher was positive but not statistically significant ($\Delta=4.2$\,pp, $p=0.146$, $d=0.34$), while LLM-as-teacher itself did not significantly outperform vanilla LLM ($\Delta=2.8$\,pp, $p=0.121$ adjusted). 
With the underlying model held constant, these results indicate a reliable advantage of AI Cosmos over conventional LLM interaction, whereas prompting the same model with a teacher persona alone did not yield a statistically detectable benefit.

\textbf{Interaction structure.}
The clearest behavioral difference emerged in the structure of interaction (Figure~\ref{legal_fig:overall}c).
Total student input did not differ across conditions (median approximately $7{,}100$ Chinese characters; $F(2,108)=0.13$, $p=0.88$), nor did session counts ($p=0.71$).
However, AI Cosmos elicited approximately $1.7\times$ as many dialogue turns as either control ($51.0\pm21.1$ vs.\ $29.6\pm13.2$ and $29.2\pm13.2$; $F(2,108)=21.8$, $p=1.1\times10^{-8}$, $\eta^2=0.288$; both pairwise $p<10^{-5}$; $g=1.23$ $[0.79,1.76]$).
Correspondingly, students produced approximately $40\%$ fewer characters per turn ($162$ vs.\ $259$/$276$; $F(2,108)=7.66$, $p<0.001$; AI Cosmos vs.\ Vanilla LLM $p\approx0.001$, $g=-0.80$).
Thus, despite similar total input volume, AI Cosmos produced a substantially more iterative, turn-rich interaction pattern.
Raw interaction volume itself was only weakly associated with learning gain (rounds: $r=0.16$, $p=0.09$; characters: $r=0.07$, n.s.), and the association between dialogue turns and gain was further attenuated after controlling for condition ($r=0.06$, n.s.).
These results suggest that the observed advantage cannot be explained simply by greater interaction volume and is instead consistent with differences in how participation was structured.

\textbf{Durability.}
At the four-week follow-up, the descriptive ordering observed immediately after learning was preserved.
Relative to pre-test, the AI Cosmos subsample showed the largest retained gain ($+18.9$\,pp), followed by LLM-as-Teacher ($+16.8$\,pp) and Vanilla LLM ($+14.7$\,pp; Figure~\ref{legal_fig:overall}d).
The AI Cosmos--Vanilla LLM difference remained positive ($\Delta=4.2$\,pp, $d=0.34$) but was smaller than the immediate effect.
With only $18$ participants per condition, however, the follow-up analysis was underpowered and yielded no significant between-group difference ($F(2,51)=0.48$, $p=0.62$).
Forgetting from the immediate to delayed post-test also did not differ significantly across conditions ($p=0.43$).
The follow-up therefore provides directionally consistent but inconclusive evidence regarding retention, and a larger delayed-test sample is needed to determine whether the immediate advantage persists over time.

\textbf{Structured application versus open argumentation.}
Decomposing the learning gains by item type suggested that the overall advantage was concentrated in the objective, scenario-based items
(Figure~\ref{legal_fig:type}).
Gains on these items were $18.2$, $21.2$, and $26.5$\,pp for Vanilla LLM, LLM-as-Teacher, and AI Cosmos ($F(2,108)=2.89$, $p=0.060$, $\eta^2=0.051$).
AI Cosmos exceeded Vanilla LLM by $8.3$\,pp ($95\%$ CI $[1.8,14.9]$, $d=0.56$, $p=0.018$; ANCOVA-adjusted $p=0.038$).
By contrast, gains on open-response items were smaller and did not differ across conditions ($5.6$--$7.8$\,pp; $F(2,108)=0.23$, $p=0.80$; AI Cosmos vs.\ Vanilla LLM $d=0.13$, n.s.).
A mixed-design ANOVA showed a strong main effect of item type ($F(1,108)=55.6$, $p<0.001$, $\eta_p^2=0.34$) but no condition-by-item-type interaction ($F(2,108)=0.95$, $p=0.39$).
Accordingly, the apparent concentration of the AI Cosmos advantage in structured application items should be interpreted descriptively rather than as evidence of moderation by item type.
The present study thus provides stronger evidence for gains in applying legal doctrine to structured fact patterns than for improvements in extended written argumentation.

\begin{figure*}[t]
\centering
    \includegraphics[width=\linewidth]{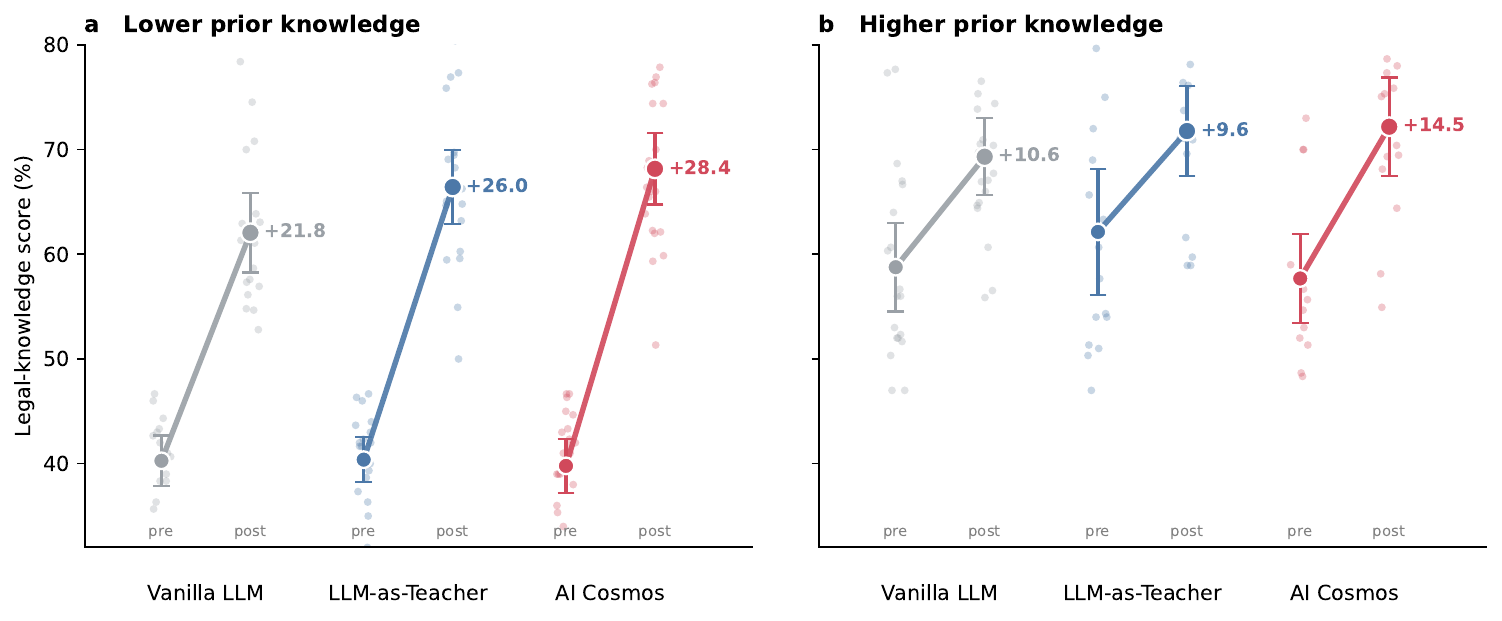}
    \caption{\textbf{AI Cosmos shows the largest learning gains across both aptitude strata.}
    Participants were split at the median pre-test score ($46.7$) into \textbf{(a) lower-} and \textbf{(b) higher-prior-knowledge strata}. Translucent points are individual pre/post scores; filled markers are group means\,$\pm$\,$95\%$\,CI. The slope of each connecting line is the mean learning gain (annotated). Within each stratum, the arms were statistically equivalent at pre-test ($p=0.92$ and $p=0.39$), yet AI Cosmos gained most in both ($+28.4$ vs.\ $+26.0$/$+21.8$\,pp, lower; $+14.5$ vs.\ $+9.6$/$+10.6$\,pp, higher). As expected given their greater headroom, lower-prior-knowledge learners gained more across all arms (aptitude main effect $F(1,105)=46.8$, $p<10^{-4}$); the group\,$\times$\,aptitude interaction was not significant ($F(2,105)=0.55$, $p=0.58$). Because the arms were baseline-equivalent within each stratum, the consistent advantage of AI Cosmos indicates the overall effect is not a statistical artifact of its slightly lower aggregate starting point.}
    \label{legal_fig:aptitude}
    \vspace{-2mm}
\end{figure*}

\textbf{Robustness across learners.}
Exploratory subgroup analyses examined whether the AI Cosmos advantage was restricted to particular learner profiles.
Across median splits on prior knowledge and AI literacy, the AI Cosmos--Vanilla LLM effect remained directionally positive, although the small subgroup sizes ($15$--$22$ participants per condition) produced wide confidence intervals and no multiplicity correction was applied
(Figure~\ref{legal_fig:forest}).
These analyses are therefore treated as consistency checks rather than tests of moderation.
A complementary split at the median pre-test score ($46.7$) showed the same descriptive ordering in both prior-knowledge strata.
Among lower-scoring learners, gains were $+28.4$, $+26.0$, and $+21.8$\,pp across the three conditions, whereas among higher-scoring learners they were $+14.5$, $+9.6$, and $+10.6$\,pp (Figure~\ref{legal_fig:aptitude}).
Lower-pre-test learners gained more overall ($F(1,105)=46.8$, $p<10^{-4}$), but the condition-by-prior-knowledge interaction was not significant ($F(2,105)=0.55$, $p=0.58$).
Together with the pre-test-adjusted ANCOVA, these results argue against the overall AI Cosmos effect being attributable solely to its slightly lower baseline performance.

\textbf{Self-efficacy and cognitive load.}
Self-efficacy gains were descriptively largest for AI Cosmos ($+5.2$ vs.\ $+4.8$ and $+3.9$ GSES points; $d=0.35$ vs.\ Vanilla LLM), but the between-group difference was not significant ($F(2,80)=1.02$, $p=0.37$).
Perceived workload also did not differ significantly across conditions (NASA-TLX: $19.4$, $18.2$, $17.5$; $F(2,78)=2.31$, $p=0.11$).
Thus, we found no evidence that the larger learning gains and more iterative interaction observed with AI Cosmos were accompanied by greater perceived workload.

\subsection{Human Study: An Instrumented Workbench for Skill Acquisition in Competitive Programming}
\label{sec:exp_algo}

Whereas the bar-examination study (\S\ref{sec:exp_legal}) primarily examined scenario-based social interaction, this study focuses on the \emph{operational} dimension: the executable workbench. 
Competitive programming provides a demanding test of procedural competence, because success requires learners not only to understand an algorithm but also to implement, execute, and debug it as a correct and efficient program. 
We thus examine whether integrating an executable coding workbench into the learning interaction improves the translation of conceptual understanding into implementation skill, and whether the effect varies with prior aptitude.

\subsubsection{Study Design and Measures}

\textbf{Design and participants.} 
We conducted a preregistered, three-arm randomized controlled study stratified by two aptitude-defined curriculum tracks. 
From the 140 students recruited, 100 completed the diagnostic pre-test, and 54 completed the full multi-session protocol and constituted the analysis sample ($93\%$ male; mean age $18.9$ years; $89\%$ first-year undergraduates). 
Participants were assigned to one of three learning conditions: 
(i) \emph{Vanilla LLM}, a unconstrained conventional request--response assistant; 
(ii) \emph{LLM-as-Teacher} baseline; and 
(iii) \emph{AI Cosmos}, which used the same underlying model within a configured scenario equipped with an executable coding workbench for implementation, execution, and iterative debugging. 
Both LLM-as-Teacher and AI Cosmos conditions followed the same restriction against supplying complete solutions, providing a focused comparison of the additional operational affordances available in AI Cosmos. 
The Vanilla LLM condition did not impose this restriction. 
All three conditions used the same base model, which was independently verified to be capable of solving the experimental tasks. 
The diagnostic pre-test was designed to align training content with prior mastery.
Participants who had not yet mastered search and dynamic programming entered the \emph{basic-algorithm track}, whereas those who had mastered these topics but not the advanced material entered the \emph{data-structure track}, which covered segment trees and leftist heaps. 
Participants already proficient in both sets of material did not proceed to the training phase. 
Within each track, participants were randomized across the three conditions ($n=19$ AI Cosmos, $18$ LLM-as-Teacher, and $17$ Vanilla LLM overall; cell sizes $7$--$11$). 
Thus, the tracks determined \emph{what} participants studied, whereas the experimental conditions determined \emph{how} they studied it. We observed no statistically significant between-condition differences in measured baseline characteristics, including pre-test performance within and across tracks and track composition (all $p\geq0.20$; Table~\ref{alg_tab:alg-baseline} in Appendix \ref{app_sec:exp_algo}; Figure~\ref{alg_fig:alg_primary}a).

\textbf{Procedure and measures.} 
Each participant completed two training sessions organized around a fixed three-stage practice cycle:
(1) \emph{Learn}: participants studied one problem interactively with their assigned system and submitted a solution to the online judge (OJ). 
(2) \emph{Self-check}: they attempted a second problem targeting the same algorithmic concept, with assistance available only when needed. (3) \emph{Quiz}: they independently solved a third problem that tested the same concept under a different problem framing. 
Advancement required a full-score OJ submission on all three problems. 
One week later, participants completed a contest-style post-test containing two previously unseen problems that required variants or combinations of the trained algorithms. 
The post-test thus assessed generalization of the trained implementation skills to novel problems. 
To minimize benchmark contamination and memorization, all problems used in the study were newly authored by two competitive-programming experts with NOI/ACM gold-medal credentials and had no online counterparts. 
The base model used in all three conditions was separately verified to be capable of solving each task, ensuring that task difficulty did not exceed the model's underlying problem-solving capability. 
Because conventional OJ scores provide limited diagnostic information about partially correct implementations, the two problem authors also rescored all submitted source code while blind to experimental condition. 
They first established a common error codebook covering compilation failures, incomplete algorithmic structures, invariant-maintenance errors, boundary and memory faults, complexity or termination failures, and degeneration to brute-force solutions (Appendix~\ref{app:alg-codebook}). 
These error categories were mapped to standardized deductions on a five-level mastery scale (G1--G5), anchored in the completeness and correctness of the target algorithm's implementation. 
The experts independently scored each submission, and their scores were then averaged. 
Expert scores were strongly correlated with OJ scores (Pearson $r=0.92$--$0.96$ across assessment phases) but captured partial algorithmic competence that binary test-case outcomes could miss. 
At post-test, expert rescoring increased the scores of 33 submissions, decreased none, and produced a mean adjustment of $+10.8$ points. We report three complementary post-test outcomes: 
(i) continuous expert-rescored performance ($0$--$100$); 
(ii) \emph{criterion mastery}, defined as grade G4 or higher, corresponding to a substantially complete and correct implementation of the target algorithm; and 
(iii) \emph{score--mastery dissociation}, capturing cases in which a submission earns partial credit without demonstrating criterion-level implementation of the target algorithm.

\textbf{Statistical analysis.} 
We report condition means and standard deviations and examine unadjusted between-condition differences using one-way ANOVA, with the Kruskal--Wallis test as a nonparametric robustness check. 
The primary between-group inference uses ANCOVA on post-test score, adjusting for pre-test performance and curriculum track. 
For complementary unadjusted pairwise comparisons, we report Welch's $t$ tests and bias-corrected Hedges' $g$ with bootstrap $95\%$ confidence intervals. 
Given the modest per-condition sample sizes ($n=17$--$19$) and the bounded or zero-inflated distributions of several outcomes, we additionally report permutation-based $p$-values and Mann--Whitney $U$ tests where appropriate. 
Criterion-mastery proportions are compared using Fisher's exact test, whereas participant-level mastery counts are compared using the Mann--Whitney $U$ test. 
Score-based results are interpreted jointly with criterion mastery, which captures algorithm-specific implementation competence not fully reflected by OJ performance alone. 
Track-stratified and submission-level analyses are exploratory and are not adjusted for multiplicity. 
All statistical tests are two-sided with $\alpha=0.05$.

\begin{figure*}[t]
    \centering
    \includegraphics[width=0.94\textwidth]{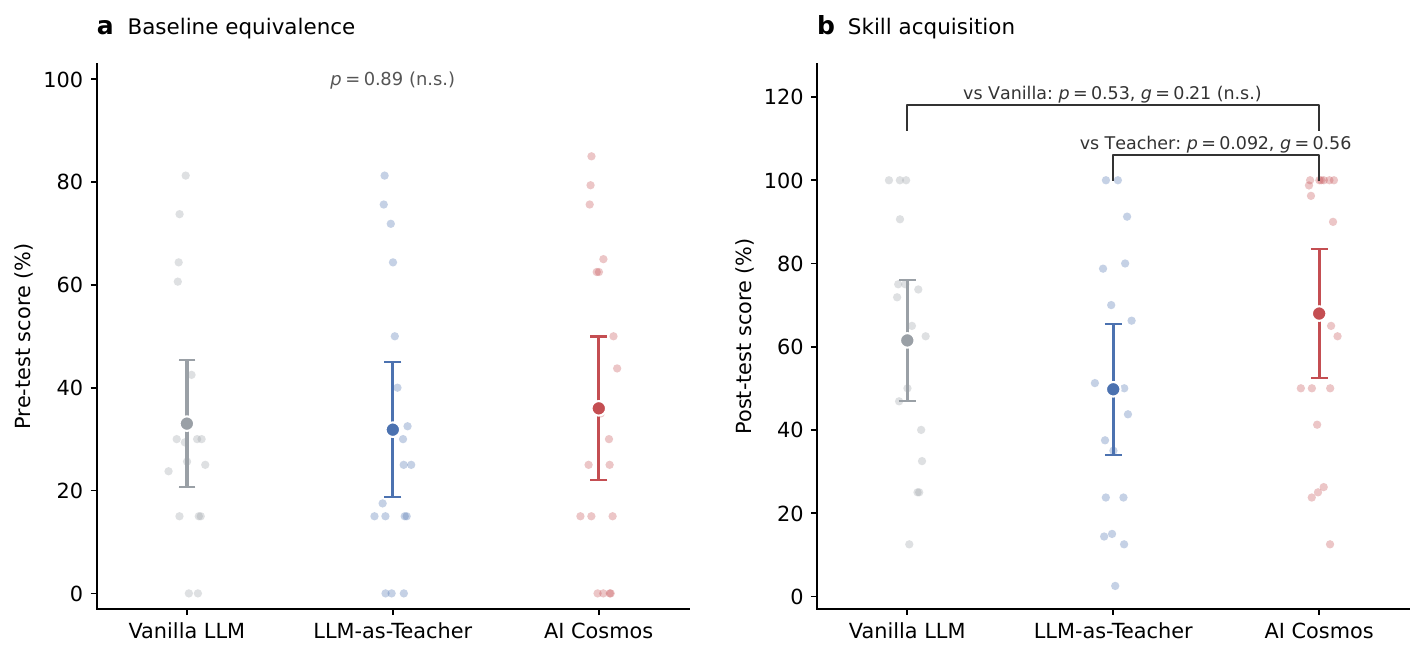}
    \caption{Results of pre- and post-test. Three arms ($n=17$--$19$) shared a single base model; both AI Cosmos and the teacher baseline withheld complete code, whereas the vanilla assistant was unconstrained.
    \textbf{(a) Pre-test scores} (dots represent individuals; markers denote mean\,$\pm$\,$95\%$ CI). The arms were statistically equivalent at baseline ($F(2,51)=0.12$, $p=0.89$).
    \textbf{(b) Post-test scores}. AI Cosmos outperformed the teacher baseline by a near-significant margin ($\Delta=18.2$ points, $g=0.56$; Welch $p=0.092$) and performed comparably to the unconstrained vanilla assistant ($g=0.21$, n.s.). The LLM-as-Teacher baseline was the weakest arm overall. Given the sample size, the omnibus test was not significant ($F(2,51)=1.65$, $p=0.20$).}
    \label{alg_fig:alg_primary}
    \vspace{-3mm}
\end{figure*}

\subsubsection{Results}

\textbf{Skill acquisition.}
Post-test scores showed a descriptive advantage for AI Cosmos ($68.0\pm32.3$), followed by the vanilla LLM ($61.5\pm28.1$) and LLM-as-Teacher ($49.8\pm31.7$; Figure~\ref{alg_fig:alg_primary}b).
The omnibus difference was not statistically significant at this sample size ($F(2,51)=1.65$, $p=0.20$, $\eta^2=0.061$; Kruskal--Wallis $p=0.22$), and the ANCOVA group effect after adjusting for pre-test score and training track showed a similar pattern ($F(2,49)=1.89$, $p=0.16$, partial $\eta^2=0.072$).
Despite the limited statistical power, the pairwise estimates reveal a consistent architectural advantage. 
AI Cosmos outperformed the LLM-as-Teacher baseline by $18.2$ points, yielding a medium effect size (Welch $t(35.0)=1.73$, $p=0.092$; permutation $p=0.091$; Mann--Whitney $p=0.096$; Hedges' $g=0.56$, $95\%$ CI $[-0.06,1.30]$).
In contrast, the LLM-as-Teacher condition produced the lowest numerical performance, scoring $11.7$ points below the vanilla assistant ($g=-0.38$).
Within-condition analyses further show that relative to their respective pre-tests, participants using AI Cosmos and vanilla LLM both improved significantly on trained material (paired $t$-tests: $p=0.0004$ and $p=0.0008$, respectively; $d_z=1.00$), whereas the LLM-as-Teacher condition showed only a non-significant increase ($p=0.12$, $d_z=0.39$).
Together, these results suggest that integrating guidance with an executable environment provides a stronger pathway from conceptual explanation to procedural programming ability than dialogue-based guidance alone.

\begin{figure*}[t]
    \centering
    \includegraphics[width=0.94\textwidth]{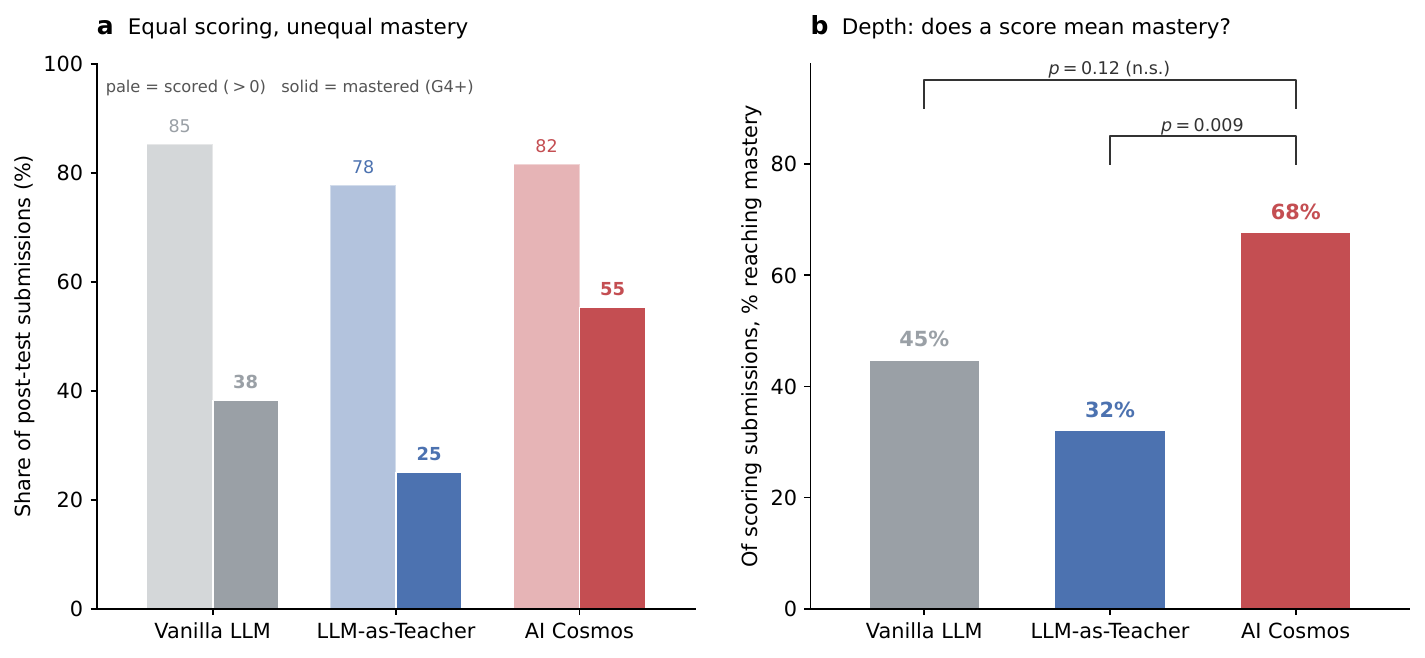}
    \caption{Post-test submission outcomes by arm.
    \textbf{(a) Rates of scored submissions ($>0$; pale bars) versus criterion mastery (grade $\geq$G4; solid bars).} AI Cosmos and the vanilla assistant earned partial credit at nearly identical rates ($82\%$ vs.\ $85\%$) but diverged sharply on true mastery ($55\%$ vs.\ $38\%$), explaining their converging mean scores despite disparate learning outcomes.
    \textbf{(b) Proportion of scored submissions achieving genuine mastery.} Among credit-earning submissions, $68\%$ under AI Cosmos reached mastery, compared to $45\%$ for vanilla and $32\%$ for the LLM-as-Teacher baseline (Fisher's exact: AI Cosmos vs.\ LLM-as-Teacher $p=0.009$; vs.\ vanilla $p=0.12$). Equivalently, AI Cosmos roughly halved the rate of hollow submissions---code that secures points without actually implementing the target algorithm.}
    \label{alg_fig:alg_dissociation}
    \vspace{-3mm}
\end{figure*}

\textbf{Equal scores, unequal mastery.}
The similar aggregate scores of AI Cosmos and vanilla LLM conceal a more fundamental difference in what learners actually achieved (Figure~\ref{alg_fig:alg_dissociation}).
Both conditions produced successful online-judge submissions at comparable rates ($82\%$ vs.\ $85\%$), explaining their similar raw scores.
Yet, automated judging primarily measures whether a program passes test cases, not whether the learner has acquired the intended algorithmic skill.
Among submissions receiving positive OJ credit, AI Cosmos achieved criterion mastery (grade $\geq$G4) in $68\%$ of cases, compared with $45\%$ for vanilla LLM and $32\%$ for LLM-as-Teacher (AI Cosmos vs.\ LLM-as-Teacher, Fisher's exact $p=0.009$; vs.\ vanilla, $p=0.12$). 
Conversely, non-mastery passing solutions, i.e., submissions that obtained test credit without substantially implementing the target algorithm, were less frequent under AI Cosmos (26\%) than under vanilla LLM (47\%) or LLM-as-Teacher (53\%; AI Cosmos vs.\ LLM-as-Teacher, $p=0.031$).
The results reveal a central limitation of score-based evaluation in AI-assisted programming: passing tests and acquiring transferable procedural knowledge are not identical outcomes.
While a conventional assistant may help learners produce partially successful solutions, AI Cosmos's executable workbench encourages learners to construct, test, and repair their own implementations, resulting in a higher proportion of solutions that reflect genuine algorithmic understanding.
Thus, the primary advantage of AI Cosmos is not merely higher task performance, but a closer alignment between measured performance and underlying competence.

\begin{figure*}[t]
    \centering
    \includegraphics[width=0.96\textwidth]{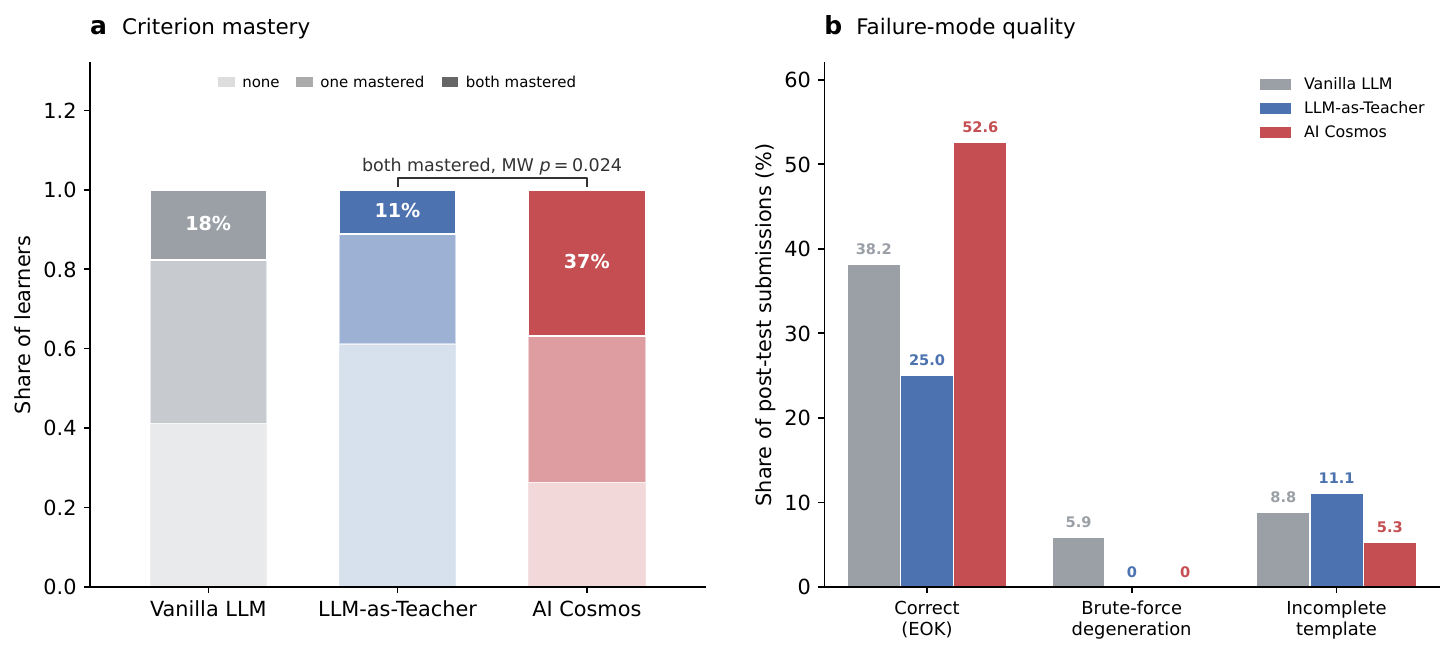}
    \caption{Criterion mastery and failure modes.
    \textbf{(a) Proportion of learners achieving criterion mastery} (grade $\geq$G4) on $0$, $1$, or $2$ post-test problems. AI Cosmos learners mastered both problems (darkest band) at more than double the rate of controls ($36.8\%$ vs.\ $17.6\%$ vanilla, $11.1\%$ LLM-as-Teacher), significantly outperforming the LLM-as-Teacher baseline in per-student mastery count (Mann--Whitney $p=0.024$). 
    \textbf{(b) Post-test failure-mode rates.} AI Cosmos achieved the highest flawless-implementation rate (\texttt{EOK}, $52.6\%$; Fisher vs.\ teacher $p=0.019$) and the lowest incomplete-template rate (\texttt{E02}). Brute-force degeneration (\texttt{E06}, i.e., exploiting test cases without using the intended algorithm) appeared only under the unconstrained vanilla arm ($5.9\%$), never under either code-withholding condition.}
    \label{alg_fig:alg_learner}
    \vspace{-3mm}
\end{figure*}

\textbf{Criterion mastery.}
Because automated scores can obscure differences in solution quality, we next examined criterion mastery: whether learners produced substantially complete and correct implementations of the target algorithms (grade $\geq$G4).
This measure captures a stricter form of procedural competence than passing test cases alone (Figure~\ref{alg_fig:alg_learner}a).
AI Cosmos learners achieved criterion mastery on an average of $1.11$ out of the two post-test problems, compared with $0.76$ for vanilla LLM and $0.50$ for LLM-as-Teacher.
The difference between AI Cosmos and LLM-as-Teacher was significant (Mann--Whitney $p=0.024$).
Moreover, $36.8\%$ of AI Cosmos learners achieved mastery on both problems, compared with $17.6\%$ for vanilla LLM and $11.1\%$ for LLM-as-Teacher.
The expert-derived error taxonomy further reveals how this advantage manifested in learners' code (Figure~\ref{alg_fig:alg_learner}b).
AI Cosmos produced the highest proportion of flawless implementations (\texttt{EOK}: $52.6\%$ vs.\ $38.2\%$ for vanilla LLM and $25.0\%$ for LLM-as-Teacher; Fisher's exact test, AI Cosmos vs.\ LLM-as-Teacher, $p=0.019$), and the lowest proportion of incomplete algorithmic structures (\texttt{E02}: $5.3\%$ vs.\ $8.8$--$11.1\%$ in the control conditions).
The ordinal mastery analysis showed a similar pattern: post-test submissions were concentrated at the two extremes of the scale, corresponding to either incomplete attempts or fully correct implementations.
A proportional-odds model further indicated that AI Cosmos learners had higher odds of achieving a stronger mastery level than LLM-as-Teacher learners (odds ratio $=2.59$, $95\%$ CI $[1.10,6.07]$, $p=0.029$).
Together, these results suggest that AI Cosmos primarily improves the quality of acquired programming skill rather than merely increasing the likelihood of obtaining a passing submission.

\begin{figure*}[t]
    \centering
    \includegraphics[width=\textwidth]{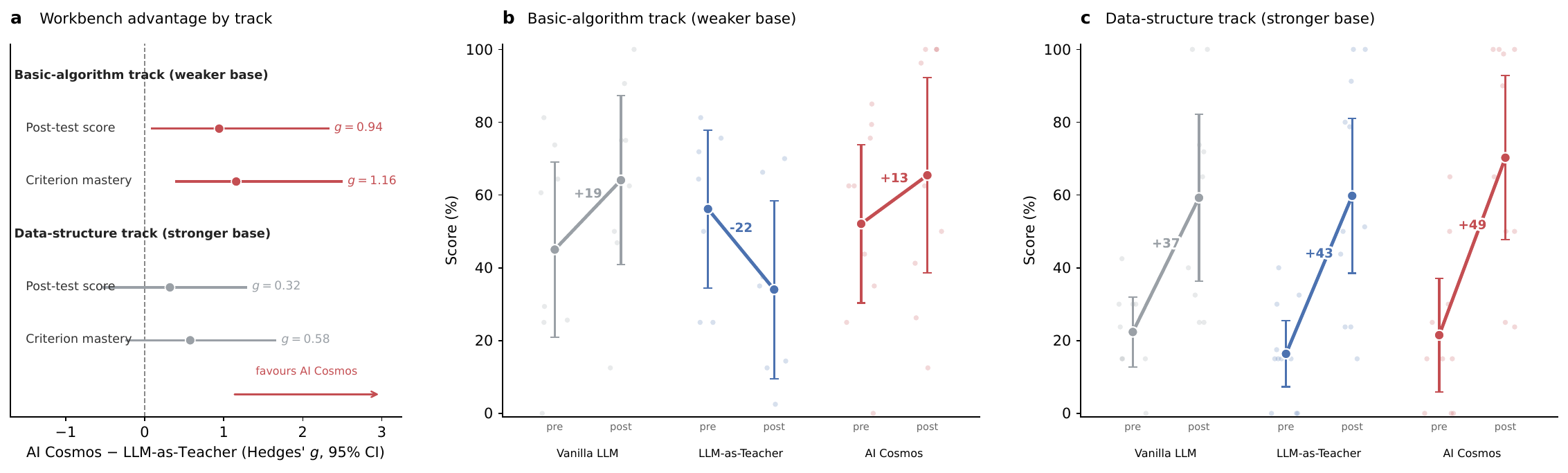}
    \caption{Differential impact of the execution workbench across learner aptitudes.
    \textbf{(a) AI Cosmos advantage over the LLM-as-Teacher baseline (Hedges' $g$, $95\%$ CI) on post-test scores and criterion mastery, by track.} The advantage is significant for mastery among novices in the basic-algorithm track (Mann--Whitney $p=0.039$). Conversely, the effect is small and statistically unresolved for the stronger data-structure track, despite the latter covering harder algorithms (arm$\,\times\,$track interaction $p=0.34$).
    \textbf{(b)\&(c) Pre- to post-test trajectories by arm within each track} (faint lines represent individuals; markers denote mean\,$\pm$\,$95\%$ CI; annotated values indicate mean change). In the basic-algorithm track, the LLM-as-Teacher baseline actually regressed from its pre-test level ($-22$ points) while AI Cosmos achieved the highest gains. In the data-structure track, all arms improved, and performance gaps narrowed.}
    \label{alg_fig:alg_workbench}
    \vspace{-3mm}
\end{figure*}

\textbf{Why the workbench matters: executable practice beyond guidance.}
The preceding analyses suggest that the advantage of AI Cosmos does not arise from providing more instructional content, but from changing how learners interact with that content via an executable environment (Figure~\ref{alg_fig:alg_workbench}).
\textbf{First}, the benefit of the workbench was largest among learners who required greater external scaffolding.
Decomposing the AI Cosmos advantage over LLM-as-Teacher by curriculum track revealed a substantial difference in the basic-algorithm track: AI Cosmos improved post-test scores by $31.4$ points, corresponding to Hedges' $g=0.94$ for scores and $g=1.16$ for criterion mastery (the latter reaching significance, Mann--Whitney $p=0.039$).
The advantage was smaller in the more advanced data-structure track ($+10.5$ points; $g=0.32$ for scores and $g=0.58$ for mastery).
Although the condition $\times$ track interaction was not significant ($p=0.34$), this pattern is consistent with the interpretation that executable environments provide greater support when learners have not yet developed stable internal models of algorithm execution.
For novice programmers, the ability to construct, execute, and debug within the same environment provides a concrete feedback loop that abstract explanations alone cannot provide.
\textbf{Second}, the programming study complements the pattern observed in the bar-examination study.
In the legal domain, where competence is expressed primarily through argumentation, additional dialogue structure produced progressively larger benefits.
Yet, in programming, competence must ultimately be expressed through executable artifacts.
Accordingly, instructional guidance without an integrated environment did not produce the same benefit as a workbench-linked scenario.
Together, the two studies suggest that effective AI learning systems must align interaction structure with the form of the target competence.
Structured dialogue supports domains where reasoning itself is the object of practice, whereas executable workbenches support domains where competence emerges through construction, testing, and revision.
AI Cosmos operationalizes this alignment by coupling guidance with the artifacts through which learners practice.

\begin{figure*}[t]
    \centering
    \includegraphics[width=\textwidth]{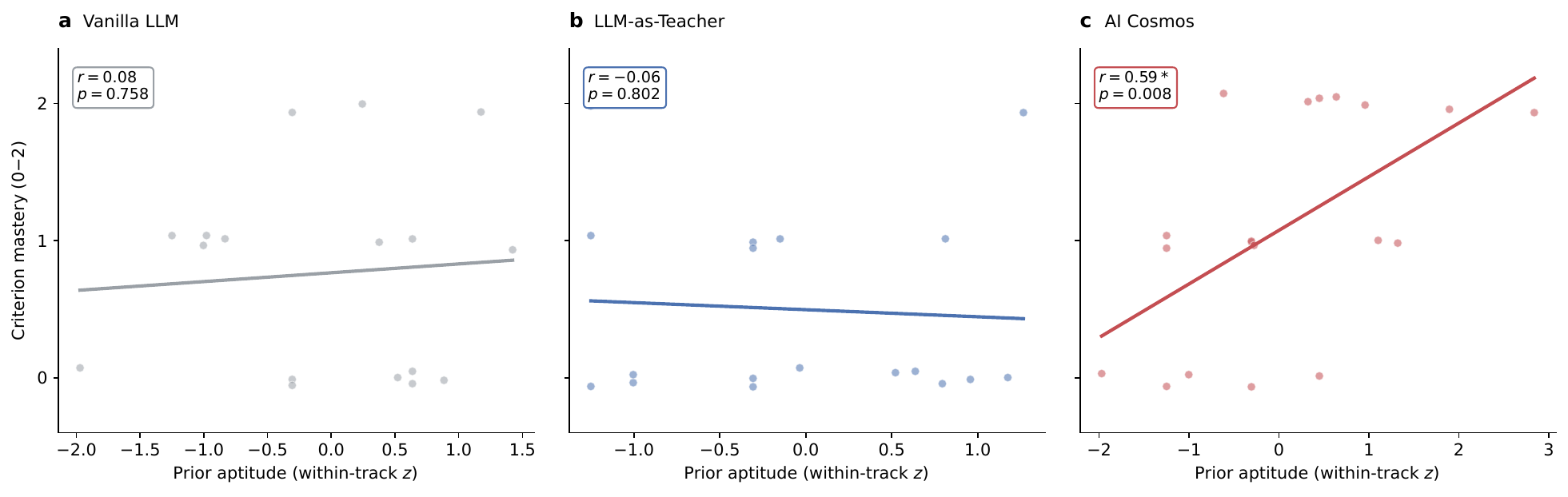}    
    \caption{Within-arm correlation between prior aptitude (within-track pre-test $z$-score) and criterion mastery. 
    Lines indicate OLS fits, and points represent vertically jittered individuals.
    \textbf{(a) Vanilla and (b) LLM-as-Teacher baselines} exhibit no predictive relationship ($r=0.08$ and $r=-0.06$, respectively; both n.s.). When an assistant can freely supply code, recorded outcomes decouple from actual learner ability.
    \textbf{(c) AI Cosmos} shows a strong, stable correlation ($r=0.59$, $p=0.008$; robust to leave-one-out resampling, $r\in[0.53,0.65]$) that significantly differs from the LLM-as-Teacher baseline (Fisher's $z$, $p=0.040$). 
    Requiring learners to construct and repair algorithms restores construct validity, making the score a faithful measure of the person.}
    \label{alg_fig:alg_validity}
    \vspace{-3mm}
\end{figure*}

\textbf{Validity: whether performance reflects learner capability.}
Finally, we conducted an exploratory construct-validity analysis examining whether observed programming performance remained associated with learners' prior ability (Figure~\ref{alg_fig:alg_validity}).
Within the AI Cosmos condition, pre-test standing was positively correlated with subsequent criterion mastery (Pearson $r=0.59$, $p=0.008$; Spearman $\rho=0.57$), and this relationship remained stable under leave-one-out resampling ($r\in[0.53,0.65]$).
In contrast, the corresponding correlations were near zero in both control conditions (vanilla LLM: $r=0.08$, $p=0.76$; LLM-as-Teacher: $r=-0.06$, $p=0.80$).
The difference between AI Cosmos and LLM-as-Teacher was significant (Fisher's $z$, $p=0.040$), with a similar but non-significant trend against vanilla LLM ($p=0.11$).
This pattern suggests that different interaction architectures may produce different relationships between measured performance and learner capability.
When learners can rely heavily on generated code, final submissions may partially reflect system assistance rather than the learner's own
implementation ability.
By requiring learners to participate in construction, execution, and debugging, AI Cosmos makes observed performance more closely coupled with the learner's demonstrated competence.
Thus, the findings highlight an important consideration for AI-assisted learning evaluation: improving performance is insufficient if the resulting measure no longer reflects what learners can independently do.

\section{Related Work}

\textbf{AI-Assisted Learning and Pedagogically Structured Interaction}\ \ \ 
Evidence from mathematics and programming education shows that improved performance with LLM assistance does not necessarily result in better performance on subsequent unassisted assessments \citep{bastani2025generative,bassner2025less}. 
Research on LLM tutoring has investigated ways to guide students during problem solving rather than directly supply answers. 
Earlier work uses annotated tutoring dialogues \citep{mathdial} and explicit pedagogical policies to study and organize multi-turn guidance \citep{pal2024autotutor,puech2025towards}. 
More recent methods train pedagogical policies \citep{dinucu2025problem}, adapt feedback using learner-state estimates and task verification \citep{wang2025training}, and maintain instructional plans across interactions \citep{li2026planning,david2026intellicode}. 
Tutoring benchmarks further assess instructional behavior beyond answer correctness \citep{maurya2025unifying,macina2025mathtutorbench}. 
These studies demonstrate progress in adaptive instructional support. However, evaluations of tutoring quality or success on guided tasks do not by themselves establish what learners can do independently. An important question is how students' own actions during assisted problem solving affect their later performance on unfamiliar tasks without assistance.

\textbf{Human–Agent Participation in Interactive Task Environments}\ \ \ 
Agent coordination can support human participation rather than only automate task completion. 
Co-STORM allows users to guide discussions among agents and reports higher-quality exploratory outputs than its comparison systems \citep{jiang2024into}. 
In a course-based multi-agent learning environment, students exhibited distinct patterns of knowledge co-construction and co-regulation, associated with prior knowledge and learning gains \citep{hao2025mapping}. 
Participation in executable tasks also requires actions that change task state and produce feedback.
CodeAct and SWE-agent improve agent performance through execution feedback and task-specific interfaces \citep{wang2024executable,yang2024swe}, while TutorGym evaluates tutoring agents on intermediate actions in interactive learning systems and identifies substantial errors in their judgments of incorrect steps \citep{weitekamp2025tutorgym}. 
Magentic-UI and Collaborative Gym further allow users to participate during agent execution \citep{mozannar2025magentic}; the latter reports better collaborative task outcomes than autonomous execution \citep{shao2026collaborative}. 
These results establish capabilities for conversational participation and shared task execution. 
For learning applications, however, task success alone does not establish which reasoning and operations learners performed themselves.

\textbf{Knowledge Building and Reusable Practice Records}\ \ \
Learning activities can produce resources that remain useful after a task is completed. 
In knowledge-building environments, students improve shared ideas through successive contributions, while Learning by Design incorporates earlier design attempts into later inquiry \citep{scardamalia2010brief,kolodner2003problem}. 
More recent tools make learners' activity records accessible: \citet{watanabe2025trackthinkdashboard} combines programming and browsing histories and identifies distinct problem-solving patterns that are not evident from final submissions alone. 
Related research on LLM agents shows that experience can also be reused for subsequent execution. 
Agent Workflow Memory extracts reusable routines and improves web-task success \citep{wang2024agent}, G-Memory retrieves collaboration histories together with generalized insights \citep{zhang2026g}, and Skill-Pro validates procedural skills for reuse across tasks and agents \citep{mi2026skill}. 
These results establish two uses of prior activity: improving an agent's later actions and helping people examine and revise earlier work \citep{skillevolver,hierachical_skill}. 
The distinction matters for educational systems, where a reusable solution does not necessarily preserve the decisions, feedback, and revisions through which learners developed it.

\section{Conclusions}

In this paper, we introduce AI Cosmos, an agentic infrastructure for situated learning that organizes AI-assisted learning around two complementary principles: scenario-based participation and cumulative assets of practice. 
Through the Scenario–Asset Framework, AI Cosmos enables learners and role-specialized agents to participate in structured learning activities, while preserving the processes and outcomes of practice as reusable assets across scenarios.
Two preregistered controlled studies in learning legal reasoning and competitive programming demonstrate the potential of structured participation to improve learners' independent competence.

\section{Contribution}

Contributors’ names are listed in alphabetical order by first name.

\textbf{Core Contributors}

Fangming Chang, Mingyu Su, Yifan Zhu, Yuan Wang

\textbf{Contributors}

Bo Zhang, Bosi Wen, Chenhao Yuan, Chenwen Liao, Dazhi Jiang, Deyao Hong, Erle Zhu, Genrui Zhang, Guchan Li, Hongwen Gan, Huichao Bi, Jiahao Yan, Jincheng Fan, Kehan Zheng, Qinren Luo, Qiran Liu, Ruoqi Cao, Shizhi Zhang, Sihan Chen, Siyan Deng, Xuehan Zhou, Xujun Li, Yang Dang, Yanzhuo Zhu, Yu Zhang, Zheyu Chen

\textbf{Tech Leads}

Jinfeng Zhou, Yuxuan Chen

\textbf{Advisors}

Hongning Wang, Manli Li

\bibliography{aicosmos}
\bibliographystyle{unsrtnat}


\newpage

\appendix

\section{Experiments}

\subsection{Human Study: Situated Learning for Bar-Examination Preparation}
\label{app_sec:exp_legal}

\subsubsection{Participant Statistics}

\begin{table*}[t]
    \centering
    \small
    \caption{Participant statistics by arm, with tests of equivalence. Values are presented as mean\,$\pm$\,s.d..}
    \label{legal_tab:barexam-baseline}
    \resizebox{.7\textwidth}{!}{
        \begin{tabular}{lcccc}
        \toprule
         & Vanilla LLM & LLM-as-Teacher & AI Cosmos & Test ($p$) \\
         & ($n=37$) & ($n=37$) & ($n=37$) & \\
        \midrule
        Female (\%)                       & $73.0$ & $81.1$ & $86.5$ & $\chi^2,\ 0.34$ \\
        Prior legal training (\%)         & $24.3$ & $27.0$ & $21.6$ & $\chi^2,\ 0.86$ \\
        Age (years)                       & $23.2\pm2.1$ & $23.8\pm3.0$ & $23.0\pm2.0$ & $F,\ 0.41$ \\
        AI literacy (/170)                & $119.7\pm15.0$ & $119.8\pm13.8$ & $116.1\pm14.3$ & $F,\ 0.44$ \\
        Self-efficacy, GSES               & $15.1\pm4.0$ & $15.1\pm4.3$ & $13.6\pm3.7$ & $F,\ 0.27$ \\
        Pre-test, total (\%)              & $50.3\pm11.8$ & $50.4\pm13.9$ & $47.0\pm11.0$ & $F,\ 0.42$ \\
        \quad- objective items (\%)       & $48.8\pm12.9$ & $48.6\pm16.0$ & $44.3\pm12.5$ & $F,\ 0.29$ \\
        \quad- open-response items (\%)   & $56.2\pm18.4$ & $57.4\pm13.8$ & $58.1\pm15.2$ & $F,\ 0.88$ \\
        \bottomrule
        \end{tabular}
    }
\end{table*}

Detailed distribution statistics of the participants are presented in Table \ref{legal_tab:barexam-baseline}.

\subsection{Human Study: An Instrumented Workbench for Skill Acquisition in Competitive Programming}
\label{app_sec:exp_algo}

\subsubsection{Participant Statistics}

Detailed distribution statistics of the participants are presented in Table \ref{alg_tab:alg-baseline}.

\begin{table*}[t]
    \centering
    \caption{Participant characteristics by arm, with tests of equivalence. Values are mean\,$\pm$\,s.d.. Pre-test scores are within-track comparable. The two tracks were assessed on different target algorithms. \textit{Pre-test problems below criterion} reports the share of pre-test submissions an expert graded below G4 (not yet a substantially complete, correct implementation of the target algorithm).}
    \label{alg_tab:alg-baseline}
    \small
    \resizebox{.7\textwidth}{!}{
    \begin{tabular}{lcccc}
        \toprule
         & Vanilla LLM & LLM-as-Teacher & AI Cosmos & Test ($p$) \\
         & ($n=17$) & ($n=18$) & ($n=19$) & \\
        \midrule
        Female (\%)                              & 11.8 & 11.1 & 0.0  & $\chi^2$, 0.31 \\
        Age (years)                              & 18.5\,$\pm$\,0.6 & 18.7\,$\pm$\,0.7 & 19.3\,$\pm$\,1.9 & $F$, 0.20 \\
        First-year (\%)                          & 94.1 & 88.9 & 84.2 & $\chi^2$, 0.64 \\
        Basic- algorithm track (\%)              & 47.1 & 38.9 & 47.4 & $\chi^2$, 0.85 \\
        Data-structure track (\%)                & 52.9 & 61.1 & 52.6 & $\chi^2$, 0.85 \\
        Pre-test (overall score)                 & 33.0\,$\pm$\,24.0 & 31.8\,$\pm$\,26.5 & 36.0\,$\pm$\,29.0 & $F$, 0.89 \\
        \quad - basic-algorithm track (score)    & 45.0 & 56.2 & 52.1 & $F$, 0.72 \\
        \quad - data-structure track (score)     & 22.4 & 16.4 & 21.5 & $F$, 0.67 \\
        Pre-test problems below criterion (\%)   & 91.2 & 86.1 & 86.8 & $\chi^2$, 0.75 \\
        \bottomrule
    \end{tabular}
    }
    \vspace{-3mm}
\end{table*}

\subsection{Expert Re-grading Rubric: Mastery Scale and Error Codebook}
\label{app:alg-codebook}
 
To convert the coarse, test-case-based online-judge (OJ) score into a measure of genuine algorithmic competence (\S\ref{sec:exp_algo}), two NOI/ACM gold-medalist experts re-graded every submission, blind to arm, using the two-part rubric reproduced here: a five-level \emph{mastery scale} (G1--G5; Table~\ref{app_alg_tab:alg_mastery_scale}) that summarizes how completely and correctly a submission realizes the problem's \emph{target algorithm}, and an \emph{error codebook} (Table~\ref{app_alg_tab:alg_error_codebook}) that records the specific failure mode underlying that grade. Each submission is tagged in the form \texttt{[code:name]} together with a grade. Throughout, the ``target algorithm'' is the intended solution for the problem at hand (e.g., flood-fill or dynamic programming on the basic-algorithm track; a segment tree or leftist heap on the data-structure track). Two grading conventions are central and explain why the expert grade can diverge from the raw OJ score. First, a submission that merely passes test cases by means of a brute-force or otherwise non-target method is \emph{not} credited as mastering the target algorithm: it is normally graded G1, regardless of the points it earns. Second, every submission that ``looks like the target algorithm but falls short of full marks'' was re-run locally and assigned a specific failure code before being graded, so that the grade reflects a diagnosed failure mode rather than the OJ verdict alone. The \textbf{criterion-mastery} threshold used in the main text is grade~$\geq$~G4 (i.e., G4 or G5); the ``flawless-implementation'' rate (\texttt{EOK}) reported there is exactly the G5 rate.

\begin{table*}[t]
    \centering
    \caption{The five-level mastery scale used by the expert re-graders. Each grade is anchored to how completely and correctly a submission realizes the problem's target algorithm. The criterion-mastery threshold used in the main text is grade~$\geq$~G4.}
    \label{app_alg_tab:alg_mastery_scale}
    \small
    \renewcommand{\arraystretch}{1.25}
    \resizebox{\textwidth}{!}{
    \begin{tabular}{@{}c >{\raggedright\arraybackslash}p{0.235\linewidth} >{\raggedright\arraybackslash}p{0.58\linewidth}@{}}
        \toprule
        Grade & Anchor & Description \\
        \midrule
        G1 & Target algorithm not formed & Implementation direction is inconsistent with the target algorithm, or no valid algorithmic intent can be identified. A brute-force or non-target solution is graded here even when it earns partial credit. \\
        G2 & Intent present, key errors & The target algorithm is attempted, but with defects in complexity, boundary handling, memory safety, or a critical invariant. \\
        G3 & Main flow present, details unstable & The target algorithm's main framework exists but still contains key implementation errors. \\
        G4 & Framework essentially complete & Only minor, non-fundamental defects remain. \emph{(Criterion-mastery threshold.)} \\
        G5 & Correct and complete & The target algorithm is implemented correctly and passes; corresponds to error code \texttt{EOK}. \\
        \bottomrule
    \end{tabular}
    }
\end{table*}

\begin{table*}[t]
    \centering
    \caption{Error codebook. Every submission is tagged \texttt{[code:name]}; the code records the specific failure mode underlying the grade. ``Severity'' reflects how far the failure is from a correct target-algorithm implementation. Codes \texttt{E01}--\texttt{E07} and \texttt{E00} are mutually informative failure modes; \texttt{N01}/\texttt{N02} are annotations rather than errors; sub-variant suffixes (e.g., \texttt{-S0}/\texttt{-S1}) distinguish source-availability and intent sub-cases and are merged into their base code for the aggregate analysis.}
    \label{app_alg_tab:alg_error_codebook}
    \small
    \renewcommand{\arraystretch}{1.2}
    \resizebox{\textwidth}{!}{
    \begin{tabular}{@{}l >{\raggedright\arraybackslash}p{0.205\linewidth} >{\raggedright\arraybackslash}p{0.50\linewidth} c@{}}
        \toprule
        Code & Name & Definition & Severity \\
        \midrule
        \texttt{E00}    & Not submitted / incomplete   & No valid submission, or no assessable result was produced.                                  & medium \\
        \texttt{E01}    & Compilation / syntax error   & Code does not compile, or the submitted file is not executable source.                       & high \\
        \texttt{E02}    & Incomplete skeleton          & The basic structure of the target algorithm was never formed.                                & high \\
        \texttt{E03}    & Critical maintenance error   & A core invariant or key maintenance step is wrong, yielding incorrect results.               & high \\
        \texttt{E04}    & State / boundary error       & The main flow is present, but a conditional branch or boundary case is mishandled.           & medium \\
        \texttt{E05}    & Complexity / termination error & Complexity is inadequate or the termination strategy is unstable, failing some test cases. & high \\
        \texttt{E06}    & Brute-force degeneration     & A non-target, brute-force implementation degrades the complexity (time-limit exceeded).      & high \\
        \texttt{E07}    & Boundary / memory-safety error & Out-of-bounds or illegal access, stack-overflow risk, or another runtime fault.            & high \\
        \texttt{E99-S0} & Source missing               & No analyzable source code is available for the problem.                                      & medium \\
        \texttt{E99-S1} & Intent unclear               & A file exists, but no valid algorithmic intent can be determined.                            & medium \\
        \texttt{EOK}    & No error                     & The target algorithm is implemented correctly and passes (grade G5).                         & none \\
        \texttt{N01}    & Non-standard but passing     & A non-standard yet theoretically valid solution that passes all local tests (flagged separately). & note \\
        \texttt{N02}    & Non-target implementation    & The approach is inconsistent with the target algorithm; not counted as target-algorithm mastery. & note \\
        \bottomrule
    \end{tabular}
    }
\end{table*}

The two parts are coupled: the error code localizes \emph{what} went wrong, and the mastery grade places that failure on a single ordinal scale of target-algorithm competence. Thus an \texttt{E06} (brute-force) or \texttt{N02} (non-target) submission is graded G1 however many test cases it passes, an \texttt{E02} (incomplete skeleton) typically maps to G2--G3, a submission with only minor defects maps to G4, and an \texttt{EOK} submission is G5. The main-text failure-mode analysis (Figure~\ref{fig:alg-learner}b) reports the rates of the three most diagnostic outcomes on the post-test problems---\texttt{EOK} (correct), \texttt{E06} (brute-force degeneration), and \texttt{E02} (incomplete skeleton)---computed over all submissions in each arm.

\end{document}